\documentclass[sigconf]{acmart}
\usepackage{multirow}
\usepackage{booktabs}

\begin{document}

\title{SAIL: Structure-Aware Interpretable Learning for Anatomy-Aligned Post-hoc Explanations in OCT}

\author{Tienyu Chang}
\authornote{Both authors contributed equally to this research.}
\affiliation{%
  \institution{Dept. of BioHealth Informatics, Indiana University}
  \city{Indianapolis}
  \state{Indiana}
  \country{USA}
}
\email{tienchan@iu.edu}
\orcid{0000-0001-7230-6226}

\author{Tianhao Li}
\authornotemark[1]
\affiliation{%
  \institution{School of Information, University of Texas at Austin}
  \city{Austin}
  \state{Texas}
  \country{USA}
}
\email{tianhao@utexas.edu}

\author{Ruogu Fang}
\affiliation{%
  \institution{Dept. of Biomedical Engineering, University of Florida}
  \city{Gainesville}
  \state{Florida}
  \country{USA}
}
\email{ruogu.fang@bme.ufl.edu}

\author{Jiang Bian}
\affiliation{%
  \institution{Dept. of Biostatistics and Health Data Science, Indiana University School of Medicine}
  \institution{Regenstreif Institute}
  \city{Indianapolis}
  \state{Indiana}
  \country{USA}
}
\email{bianji@iu.edu}

\author{Yu Huang}
\affiliation{%
  \institution{Dept. of Biostatistics and Health Data Science, Indiana University School of Medicine}
  \institution{Regenstreif Institute}
  \city{Indianapolis}
  \state{Indiana}
  \country{USA}
}
\email{yh60@iu.edu}

\renewcommand{\shortauthors}{TienYu et al.}

\begin{abstract}

Optical coherence tomography (OCT), a commonly used retinal imaging modality, plays a central role in retinal disease diagnosis by providing high-resolution visualization of retinal layers. While deep learning (DL) has achieved expert-level accuracy in OCT-based retinal disease detection, its "black box" nature poses challenges for clinical adoption, where explainability is essential for clinical trust and regulatory approval. Existing post-hoc explainable AI (XAI) methods often struggle to delineate fine-grained lesion structures, respect anatomical boundaries, or suppress noise, limiting the trustworthiness of their explanations. 

To bridge these gaps, we propose a Structure-Aware Interpretable Learning (SAIL) framework that integrates retinal anatomical priors at the representation level and couples them with semantic features via a fusion design. Without modifying standard post-hoc explainability methods, this representation yields sharper and more anatomically aligned attribution maps. Comprehensive experiments on diverse OCT datasets demonstrate that our structure-aware method consistently enhances interpretability, producing clinically meaningful and anatomy-aware explanations. Ablation studies further show that strong interpretability requires both structural priors and semantic features, and that properly fusing the two is critical to achieve the best explanation quality. Together, these results highlight structure-aware representations as a key step toward reliable explainability in OCT.

\end{abstract}




\maketitle

\section{Introduction}
\label{sec:intro}

Vision impairment and retinal diseases continue to impose a growing global health burden, underscoring the importance of early diagnosis and adequate follow-up \cite{burton2021lancet,eye_image,li2023global,zhang2024global,zhou2023visual}. Retinal imaging is central to ophthalmic diagnosis and monitoring \cite{eye_image}. Optical coherence tomography (OCT) has become the clinical workhorse and a gold standard of initial assessment in many conditions (e.g., age-related macular degeneration [AMD], diabetic macular edema [DME], Epiretinal Membrane [ERM]) \cite{amd2,lim2025diabetic,amd1,wilkinson2003proposed}, because it offers high-resolution visualization of retinal microstructure and supports precise evaluation of retinal thickness and structural integrity \cite{oct_app,oct_app2}. Yet, the growth in OCT use has outpaced the availability of specialist capacity, motivating automated methods that can support screening and scan interpretation \cite{eye_image,eye_image2,eye_image3}. Artificial intelligence (AI), particularly deep learning (DL), offers a promising scalable approach to automate OCT analysis for accurate and efficient eye disease diagnosis \cite{oct_app,de2018clinically,grace2021investigation,lim2025diabetic,markan2020novel}.

DL has demonstrated strong performance across various retinal image diagnosis tasks, in some settings reaching expert-level accuracy \cite{dl_ophthalmology,dl_overview,dl_radiology}. Yet, high predictive accuracy alone is insufficient for clinical adoption, and limited model trustworthiness can undermine clinical reliability. Explainable AI (XAI) methods aim to bridge this gap by providing human-understandable rationales for model predictions; in retinal image analysis, they are often visualized as attribution (saliency) maps \cite{dl_XAI_review}. Several studies in OCT-based ocular disease detection have applied XAI to interpret DL models and assess whether highlighted regions align with clinically meaningful structures \cite{oct_dl2,oct_dl1,dme3}. Notably, Yoshida et al. \cite{dme3} recruited retinal specialists to assess the clinical relevance of saliency regions from XAI, identifying two known disease signs suggesting potentially novel cues. These findings underscore the potential of integrating XAI with high-performing DL models to enhance both clinical interpretability and scientific discovery.

Despite this promise, widely used XAI methods (e.g., Grad-CAM) were developed for natural images and are often applied to OCT without accounting for retinal-specific structure and artifacts. OCT exhibits strong anatomical organization (e.g., layered boundaries), modality-specific noise (e.g., speckle and shadowing), and pathology that is typically confined to particular layers rather than arbitrary regions. Yet most classification backbones are optimized for image-level prediction accuracy and do not encode such anatomy. Consequently, their post-hoc attribution maps are often diffuse, anatomy-agnostic, or artifact-sensitive, limiting their interpretability and clinical reliability. More broadly, evaluation studies in medical imaging report that popular XAI methods can fall short of expert expectations \cite{XAI_problems}, especially for small or morphologically complex findings, motivating explanations that better respect anatomy and image-formation characteristics. From a representation learning perspective, this issue reflects a mismatch between features optimized for disease discrimination and the structural priors required for anatomically faithful explanations \cite{gao2020feature,zhao2023multi}.

To address these limitations, we propose the Structure-Aware Interpretable Learning (SAIL) framework that incorporates OCT anatomical priors into the model training pipeline to maintain competitive diagnostic performance while improving anatomical alignment. The key idea is to first learn the intrinsic anatomical structural properties (e.g., retinal layer segmentation) through segmentation-based pretraining and then transfer this structural knowledge to disease diagnosis. By combining anatomical representations with diagnosis semantics, the model is encouraged to focus on clinically meaningful retinal regions while suppressing spurious background signals, yielding explanations that are more aligned with retinal anatomy. We leverage a segmentation backbone pretrained for OCT layer segmentation to produce anatomically informed features that are then adapted for disease classification. We introduce a fusion module that links classification decisions to expert-defined retinal layer structure, enabling layer-aware attribution. To evaluate explanation quality, we introduce Relevance Mass Accuracy and Relevance Rank Accuracy metrics \cite{eye_image,xai_metric,xai_interpretable} to measure where attribution mass lies within retinal tissue.

Overall, our framework effectively addresses two practical challenges in OCT-based analysis: (1) OCT B-scan images contain substantial background and spatially heterogeneous artifacts that can distract standard models and lead to spurious explanations, and (2) clinically meaningful interpretation often depends on understanding which retinal layers drive a decision. Our segmentation-based anatomical priors and encoder–decoder fusion offer clinically interpretable and quantitative explanations for model decisions with common XAI methods.

In summary, our work makes the following contributions:
\begin{enumerate}
    \item We introduce SAIL, a structure-aware learning framework that uses retinal layer segmentation supervision to learn anatomy-preserving features and injects this structural information into the classification representation at prediction, encouraging post-hoc attribution maps to align with retinal anatomy rather than artifacts.
    \item We propose a segmentation-guided fusion module that couples semantic features with anatomy-preserving structural features through a gated fusion. This produces an attribution target that jointly encodes semantics and structure, improving anatomical fidelity of attribution maps without changing the post-hoc explanation method.
    \item We propose an anatomy-aware evaluation protocol with pixel and layer-level analyses: RMA/RRA quantify retina-confined attribution, and layer-wise attribution links salient layers to established clinical evidence.
    \item We evaluate SAIL on two public benchmarks and a large-scale real-world cohort, demonstrating consistent gains in anatomy alignment and clinically plausible attribution maps under unchanged XAI methods. We further isolate the source of improvements via feature-source and fusion ablations and complement quantitative results with a qualitative analysis.
\end{enumerate}

\section{Related Work}
\label{sec:related}

\subsection{Explainable AI for Imaging}

Within XAI, most visual explanation methods produce attribution maps that highlight input regions deemed important for a model’s prediction \cite{xai_comp}. Common families include attribution, perturbation, and attention-based approaches. In attribution methods, gradient-based approaches, such as Grad-CAM \cite{gradcam}, Grad-CAM++ \cite{gradcam++}, and HiResCAM \cite{hirescam}, use backpropagation gradient information to localize discriminative regions. Relevance-propagation methods, such as Layer-wise Relevance Propagation (LRP) \cite{lrp_2016} and Concept Relevance Propagation (CRP) \cite{crp}, back-propagate relevance scores through the network to assign pixel-level attributions. 
Perturbation-based methods explain predictions by measuring output changes under controlled input modifications. For example, LIME \cite{lime} locally approximates the model with an interpretable surrogate trained on perturbed samples, and RISE \cite{rise} randomly samples binary masks to construct pixel-wise attribution maps. For attention-based models (e.g., Transformer), explanation is often derived from attention weights or combinations of attention and gradients \cite{Transformer_Explainability}. However, directly applying generic XAI methods to medical images can be problematic. In modalities such as OCT, explanations may be sensitive to acquisition artifacts and may not align with anatomical structure, particularly for small or morphologically complex findings \cite{XAI_problems_OCT}.

\subsection{Explainable AI Methods for OCT Imaging}

In OCT analysis, early explainability efforts often relied on handcrafted features \cite{OCT_XAI1,OCT_XAI2,OCT_XAI3,OCT_reflect1} paired with interpretable models (e.g., linear models) or feature-attribution tools (e.g., SHAP \cite{lundberg2017unified}). While these pipelines are able to align with clinical concepts, they typically compress spatial structure into summary features and do not directly support localized, anatomy-aware explanations. With the advent of deep learning, OCT-based disease detection has increasingly adopted convolutional neural network (CNN)- and transformer-based models with generic post-hoc XAI (e.g., Grad-CAM, LIME, or attention maps) to visualize evidence regions \cite{oct_dl2,oct_dlml,oct_dl1}. More recently, Yoshida et al. \cite{dme3} leveraged the foundation model, RETFound \cite{retfound}, for diabetic retinopathy classification and employed RELPROP \cite{Transformer_Explainability} to generate attribution maps. Such clinician-facing evaluations further confirm that attribution maps may correspond to known biomarkers and may highlight additional cues.

However, most OCT-XAI practice inherits XAI methods developed for natural images without modeling the OCT-specific structure. OCT has layered anatomy and modality-specific artifacts, and pathology is often layer-constrained; thus, generic XAI methods may yield anatomy-agnostic or artifact-driven attributions. Prior OCT-XAI work largely adopts natural-image XAI methods without injecting anatomical priors during training.

\section{Method}
\label{sec:method}

\subsection{Scientific Objective}
Retinal OCT is interpreted through layered anatomy: many markers (e.g., fluid) and pathological signs (e.g., lesion patterns) are meaningful in relation to retinal anatomy. For DL models to be scientifically and clinically useful, explanations should therefore be anatomy-resolved, and they should indicate where evidence lies in a way that respects retinal structure and be robust to OCT-specific nuisance factors (speckle, shadowing, background region correlations) \cite{shi2024retinal,wen2024concept}. Our goal is to obtain anatomy-consistent explanations. Instead of modifying XAI methods, we design the SAIL framework whose internal representation makes anatomy-aware attribution maps arise naturally when post-hoc XAI methods are applied. 

\begin{figure}[htbp]
    \centering
    \includegraphics[width=\linewidth]{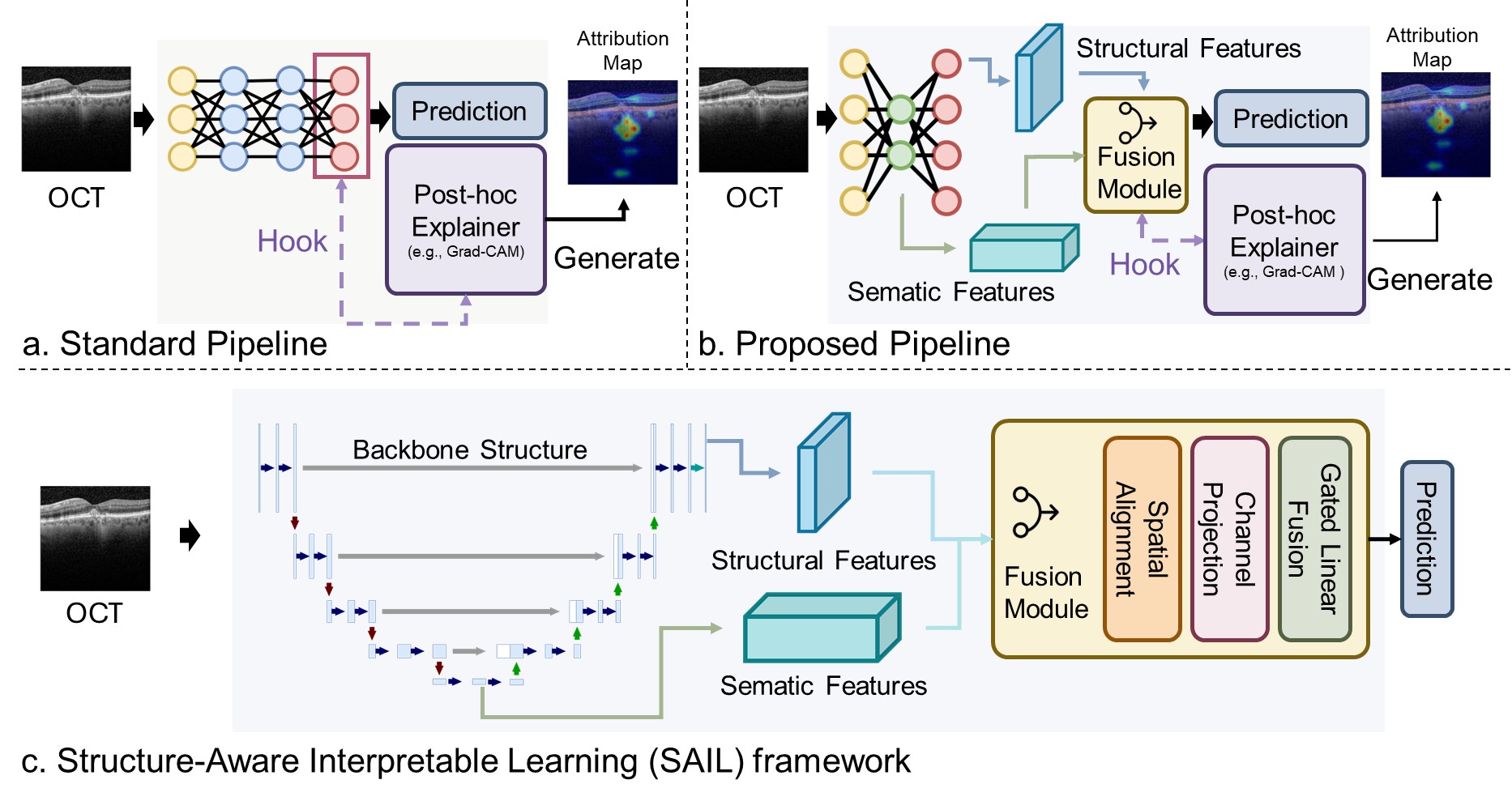}
    \caption{Overview of the proposed SAIL framework. (a) Standard model explanation pipeline using post-hoc XAI methods. (b) Proposed anatomy-informed explanation pipeline. (c) Detailed architecture of SAIL, consisting of an encoder--decoder backbone and a fusion module for structure-aware learning. }
    \label{fig:sail_framework}
\end{figure}

\textbf{Key idea.} Unlike the standard pipeline (Figure~\ref{fig:sail_framework}-a), which attaches a post-hoc XAI method to a trained backbone and typically probes a single target layer (e.g., the last layer before the prediction head), our approach injects anatomical priors directly into model representation. Concretely, we leverage semantic and structural cues already learned by the backbone (Figure~\ref{fig:sail_framework}-b) and transfer the structural information from retinal layer segmentation into the feature map used by the classifier for decision-making. This turns anatomy alignment into a property of the model representation, not just a post-hoc visualization regularization.

\subsection{Problem Setup}

Let $I \in \mathbb{R}^{H \times W}$ denote an OCT B-scan and $y \in \{1, \dots, K\}$ denote the corresponding disease label. During segmentation pretraining, we assume access to a retinal layer annotation map $S \in \{1, \dots, L\}^{H \times W}$, where $L$ denotes the number of retinal layer classes.

Given a trained model $f(\cdot)$, a post-hoc explanation method produces an attribution map $A \in \mathbb{R}^{H \times W}$ highlighting regions that contribute to the predicted class. Our goal is to learn a classifier whose internal representation is structure-aware, such that attribution maps derived from $f(\cdot)$ are spatially precise and anatomically consistent.

\subsection{SAIL Model Architecture}
We instantiate the SAIL backbone $f$ using an encoder-decoder architecture with a fusion module (Figure~\ref{fig:sail_framework}-c). The backbone follows a U-Net--style encoder--decoder architecture $f_{\text{U-Net}} = f_{\mathrm{enc}} \circ f_{\mathrm{dec}}$. 
We adopt this architecture as it is designed for pixel-wise prediction. The contracting path captures multi-scale semantic context, while the expanding path recovers fine spatial detail needed for precise anatomical boundary localization (e.g., retinal layer boundary, lesion area). Specifically, encoder features provide increasingly abstract, context-rich representations (semantic cues) as resolution decreases, while the decoder restores spatial precision through upsampling and skip-connected fusion with high-resolution encoder features, emphasizing anatomical structure (i.e., boundaries) in the final prediction.

\subsubsection{Encoder–Decoder Feature Extraction.}
Given an input OCT B-scan $I$, the encoder extracts hierarchical semantic features:
\[
F_{\mathrm{enc}} = f_{\mathrm{enc}}(I),
\]
and the decoder produces spatially detailed representations:
\[
F_{\mathrm{dec}} = f_{\mathrm{dec}}(F_{\mathrm{enc}}).
\]
We extract features from designated encoder and decoder stages: $F_{\mathrm{enc}} \in \mathbb{R}^{B \times C_e \times H_e \times W_e}$ and $F_{\mathrm{dec}} \in \mathbb{R}^{B \times C_d \times H_d \times W_d}$, where $B$ denotes the batch size, and $(C, H, W)$ represent the number of channels, height, and width of the feature representations, respectively. The actual values of encoder and decoder channels, heights, and widths are specified in the implementation details.

\subsubsection{Fusion Module.}
The encoder feature $F_{\mathrm{enc}}$ captures class discriminative semantics but is spatially coarse, whereas the segmentation pretrained decoder feature $F_{\mathrm{dec}}$ preserves retinal-layer geometry but is less class discriminative. We fuse them into $F_f$, so both prediction and Grad-CAM attributions are computed in the same structure-aware feature space.

\begin{figure}[htbp]
    \centering
    \includegraphics[width=\linewidth]{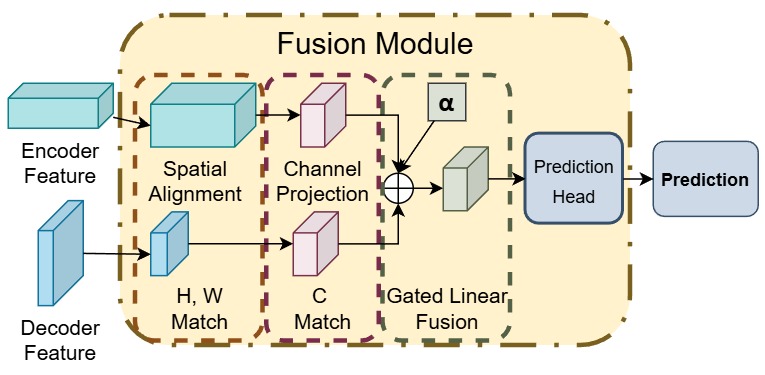}
    \caption{Encoder--decoder fusion module. Encoder feature $F_{\mathrm{enc}}$ and decoder feature $F_{\mathrm{dec}}$ are resized to a common spatial resolution ($H, W$), projected to a shared channel dimension ($C$), and fused using a gated linear combination. The fused feature $F_f$ is used for both classification and attribution.}
    \label{fig:fusion_head}
\end{figure}

\textbf{Spatial alignment}. Since encoder and decoder features have different resolutions (Figure~\ref{fig:fusion_head}), we align them to a common spatial size $(H,W)$ using a deterministic resize operator $R(\cdot)$:
\[
\tilde{F}_{\mathrm{enc}} = R(F_{\mathrm{enc}}), \qquad
\tilde{F}_{\mathrm{dec}} = R(F_{\mathrm{dec}}),
\]
where $\tilde{F}_{\mathrm{enc}} \in \mathbb{R}^{B \times C_e \times H \times W}$ and $\tilde{F}_{\mathrm{dec}} \in \mathbb{R}^{B \times C_d \times H \times W}$. In practice, we upsample $F_{\mathrm{enc}}$ using bilinear interpolation and downsample $F_{\mathrm{dec}}$ using adaptive average pooling.

\textbf{Channel projection}. The $1 \times 1$ convolutions are parameter-efficient and spatially preserving, enabling channel alignment without altering spatial structure. We then project both aligned tensors $\tilde{F}_{\mathrm{enc}},\tilde{F}_{\mathrm{dec}}$ to a shared channel dimension $C$ using $1 \times 1$ convolutions $\phi$:
\[
\hat{F}_{\mathrm{enc}} = \phi_{\mathrm{enc}}(\tilde{F}_{\mathrm{enc}}), \qquad
\hat{F}_{\mathrm{dec}} = \phi_{\mathrm{dec}}(\tilde{F}_{\mathrm{dec}}),
\]
where $\hat{F}_{\mathrm{enc}}, \hat{F}_{\mathrm{dec}} \in \mathbb{R}^{B \times C \times H \times W}$. Here, $C$, $H$, and $W$ denote the final channel number, height, and width, respectively. 

\textbf{Gated linear fusion.} General Class Activation Mapping (CAM) methods assume an approximately linear readout from a feature map to the class score (often via global pooling). Linear mixing maintains a clear decomposition between semantic and structural evidence, yields stable gradients for CAM methods, and keeps the fusion mechanism inspectable. Next, we fuse semantic and structural features using a gated linear fusion:
\[
F_f = \alpha \hat{F}_{\mathrm{enc}} + (1 - \alpha)\hat{F}_{\mathrm{dec}}, 
\qquad \alpha = \sigma(w) \in (0,1),
\]
where \(w\) is a learnable scalar, \(\sigma(\cdot)\) is the sigmoid function, and the resulting gate \(\alpha\) is shared across all spatial locations and channels. Learned jointly with the classification objective, \(\alpha\) adaptively balances class-discriminative semantic features and anatomy-preserving structural features while introducing minimal additional parameters.

\textbf{Prediction head.} Finally, we compute prediction $\hat{\mathbf{y}}$ from the fused map using a lightweight head $\psi(\cdot)$ followed by global average pooling and softmax function:
\[
\mathbf{z} = \mathrm{GAP}\bigl(\psi(F_f)\bigr), \qquad
\hat{\mathbf{y}} = \mathrm{Softmax}(\mathbf{z}).
\]

\subsection{Stage-wise Training Paradigm}

We train the model in two stages: 
(1) segmentation pretraining to learn retinal layer structure, and 
(2) classification fine-tuning using the pretrained encoder--decoder augmented with the proposed fusion module.

\subsubsection{Stage I: Segmentation Pretraining.}
This pretraining stage encourages the model to learn anatomically meaningful representations, including retinal layer boundaries and layer-specific texture patterns. We pretrain the encoder--decoder network on OCT retinal layer segmentation:
\[
\hat{S} = f_{\mathrm{seg}}\bigl( f_{\text{U-Net}}(I) \bigr),
\]
where $f_{\text{U-Net}}$ denotes the U-Net backbone and $f_{\mathrm{seg}}$ is the segmentation head producing the final output mask.

During segmentation pretraining, we optimize a hybrid segmentation loss that combines Dice loss and cross-entropy loss:
\[
\mathcal{L}_{\mathrm{seg}} = 
\mathcal{L}_{\mathrm{Dice}}(\hat{S}, S) 
+ \lambda \, \mathcal{L}_{\mathrm{CE}}(\hat{S}, S),
\]
where $S$ is the ground-truth retinal layer mask, $\mathcal{L}_{\mathrm{Dice}}$ denotes the Dice loss, $\mathcal{L}_{\mathrm{CE}}$ denotes the cross-entropy loss, and $\lambda$ controls the contribution of the cross-entropy term. 

\subsubsection{Stage II: Classification Fine-tuning.}

After segmentation pretraining, we initialize the classification backbone with the pretrained encoder and decoder weights, and attach the fusion module and classification head. Although no explicit structural regularization is imposed during this stage, structural constraints are implicitly enforced through the fused representation used for classification. The overall forward computation is given by
\[
\mathbf{z} = f_{\mathrm{head}}\bigl(
f_{\mathrm{enc}}(I),\;
f_{\mathrm{dec}}\bigl(f_{\mathrm{enc}}(I)\bigr)
\bigr), 
\qquad
\hat{\mathbf{y}} = \mathrm{Softmax}(\mathbf{z}).
\]

The model is trained using the standard cross-entropy loss:
\[
\mathcal{L}_{\mathrm{cls}} = 
- \sum_{k=1}^{K} y_k \log \hat{y}_k,
\]
where $y_k$ denotes the one-hot ground-truth label and $\hat{y}_k$ denotes the predicted probability for class $k$.

\subsection{Explanation Generation and Anatomy-Guided Evaluation}

We apply unchanged post-hoc XAI methods (i.e., Grad-CAM and related approaches) to the trained classifier. The improved localization of the resulting attribution maps arises naturally from the anatomy-informed feature representation.

When retinal layer masks $\Omega$ are available, attribution maps can be summarized in a layer-wise manner. Let $\Omega_l$ denote the set of pixels corresponding to retinal layer $l$. Given an attribution map $A \in \mathbb{R}^{H \times W}$, the layer-wise relevance mass is defined as
\[
\mathrm{m}(l) = \sum_{(i,j) \in \Omega_l} A(i,j),
\]
and the normalized relevance distribution over layers is given by
\[
\tilde{\mathrm{m}}(l) = 
\frac{\mathrm{m}(l)}
{\sum_{l' = 1}^{L} \mathrm{m}(l')}.
\]

This formulation enables quantitative evaluation of the anatomical faithfulness and consistency of explanations, while preserving standard post-hoc XAI pipelines and avoiding the need for additional supervision.

\section{Experimental Setup}
\label{sec:experiment_setup}

\subsection{Datasets}

We evaluate our proposed method and baselines on two public OCT classification benchmarks and a large real-world private UF cohort, covering both large- and small-scale settings. We also use two public datasets for retinal-layer segmentation pretraining.

\textbf{UF (private RWD cohort)}: Data are from the University of Florida (UF) Health System. We include patients who underwent eye-related procedures or ophthalmology visits involving retinal imaging between Jan 1, 2012, and Jan 1, 2024. We study four classification tasks: DME, AMD, Glaucoma, and ERM. The cohort contains 5,498 control patients (14,649 DICOM) with no retinal disease and 16,538 disease patients (115,134 DICOM). Cohort construction and detailed statistics are provided in Appendix~\ref{app:uf_process}.

\textbf{OCTDL} \cite{octdl}: OCTDL includes 1,231 AMD, 147 DME, 155 ERM, 332 normal OCT B-scans, plus other retinal conditions (e.g., RAO, RVO, VMI disease). We use DME, AMD, and ERM for disease classification in this work.

\textbf{OCT2017} \cite{oct2017}: OCT2017 contains 108,312 OCT B-scans from 4,686 patients (CNV, DME, drusen, normal) with a held-out test set of 1,000 B-scans (250 B-scans per class) from 633 patients. We use DME for disease classification in this work.

\textbf{DUKE DME} (segmentation) \cite{dukedme}: DUKE DME dataset contains 110 SD-OCT B-scans from 10 DME patients with expert annotations for 7 retinal layers, fluid, and background.

\textbf{NR206} (segmentation) \cite{nr206}: NR206 dataset contains 206 B-scans of healthy eyes (from OCTID), annotated into 8 retinal layers and background.

\subsection{Baseline Models \& XAI}

We benchmark SAIL against representative CNN and transformer models, including ResNet-50 (ResNet) \cite{resnet}, EfficientNet-B4 (EfficientNet) \cite{effnet}, Vision Transformer (ViT) \cite{vit}, and RETFound \cite{retfound}, an OCT-pretrained foundation model.

For our method, we evaluate two variants in the main experiments. \textbf{SAIL} denotes the full model described in Section~\ref{sec:method}. \textbf{SAIL-enc} is a reduced variant that only uses the segmentation-pretrained encoder features. Comparing SAIL and SAIL-enc isolates the contribution of segmentation-guided fusion beyond structure-aware encoder representations alone.

For explainability, we focus on gradient-based CAM variants that are widely used in medical imaging. We adopt Grad-CAM \cite{gradcam}, Grad-CAM++ \cite{gradcam++}, and HiResCAM \cite{hirescam} as our testing XAI methods. We do not include perturbation-based explanation methods (e.g., RISE \cite{rise}), as they rely primarily on input perturbations rather than model-internal representations and are therefore less suitable for analyzing the effects of structure-aware feature design.

\subsection{Implementation Details}

\subsubsection{Segmentation Pretraining.}
We pretrain a retinal-layer segmentation model on Duke DME and NR206 datasets using a U-Net (ResNet-50 encoder) with data augmentations. We follow the original dataset training and validation splits and report Dice and Intersection over Union (IoU) score. The full preprocessing, training, and experimental results are in Appendix~\ref{app:segmentation}.

\subsubsection{Disease Classification.}
We train all of the models under a unified setup. RETFound uses the official public checkpoint; other baselines use ImageNet initialization. SAIL is initialized from the segmentation checkpoint and fine-tuned in different settings. The fusion height, width, and channel are 32, 32, and 8.
We use the middle B-scan from each OCT series as input. All inputs are resized to 224×224 and ImageNet-normalized. Models are trained for 50 epochs with AdamW (learning rate $5\times{10}^{-4}$, weight decay 0.05) and cosine schedule with 10-epoch warmup. We select the best checkpoint by means of F1, AUROC, and Cohen’s kappa on validation as RETFound, then evaluate on a held-out test set. The same test set is used for XAI evaluation.


\subsection{Evaluation Metrics}

We report standard classification metrics to provide each model's performance before analyzing explanations. We use AUROC as the primary metric. Additionally, we report AUPRC, accuracy, precision, recall (sensitivity), F1, and Cohen’s kappa in Appendix~\ref{app:full_class_perf}

To quantify the faithfulness of attribution maps, we use insertion and deletion AUC \cite{rise}. For each test image, pixels are perturbed in descending attribution score order, and the model confidence curve is summarized by AUC. In deletion, it progressively removes top-attribution pixels; lower AUC is better. In insertion, it starts from a blurred baseline and progressively reveals top-attribution pixels; higher AUC is better.

For explainability aligned with domain knowledge, we assess whether the attribution lies within retinal tissue using OCTExplorer layer segmentation ~\cite{octexplorer1,octexplorer2,octexplorer3} as the reference retinal-layer masks. We then compute Relevance Mass Accuracy (RMA) and Relevance Rank Accuracy (RRA) ~\cite{xai_metric}. Let $A$ denote the attribution map, $\Omega$ denote the set of all pixels, and $\mathcal{G}\subseteq\Omega$ denote the set of retinal pixels obtained from OCTExplorer (ground-truth). We define
\begin{equation}
\mathrm{RMA}
= \frac{\sum_{p \in \mathcal{G}} A_p}{\sum_{p \in \Omega} A_p},
\qquad
\mathrm{RRA}
= \frac{\left|\mathcal{P}_{\mathrm{top}K} \cap \mathcal{G}\right|}{\left|\mathcal{G}\right|},
\end{equation}
where $A_p$ is the relevance value at pixel $p$, and $\mathcal{P}_{\mathrm{top}K}$ is the set of the top-$K$ pixels ranked by relevance value (largest $A_p$), with $K = |\mathcal{G}|$.
RMA captures the fraction of total relevance mass inside the retinal tissue, while RRA measures how concentrated the highest-relevance pixels are within the retinal tissue.

\subsection{Qualitative Analysis}

To assess whether attribution maps support downstream clinical interpretation, we conduct a qualitative analysis of our method. We use two complementary components: targeted case studies and layer attribution studies.

\textbf{Case Studies.} For each task, we sample one representative OCT B-scan and inspect attribution maps for three criteria: (1) localization within retinal tissue (vs. background/artifacts) and (2) emphasis on clinically plausible structures (e.g., fluid or layer).

\textbf{Layer Attribution Studies.} To quantify whether explanations concentrate on clinically relevant retinal layers, we aggregate attribution scores within OCTExplorer-derived layer masks to obtain per-layer relevance scores. We then identify the top-3 layers by relevance mass, report the most frequent top-ranked layers (dominant layers) per task, and compute the Top-3 ratio:
\begin{equation}
\mathrm{Top\text{-}3\ Ratio}=\frac{m_{\pi(1)} + m_{\pi(2)} + m_{\pi(3)}}{\sum_{l=1}^{L} m_l},
\end{equation}
where $m_l$ is the relevance mass for layer $l \in \{1,\dots,L\}$, and $m_{\pi(i)}$ denotes the $i$-th largest value among $\{m_l\}_{l=1}^{L}$ (i.e., $m_{\pi(1)} \ge m_{\pi(2)} \ge m_{\pi(3)}$).

\section{Experimental Results}
\label{sec:results}
We evaluate SAIL on two public OCT datasets (OCTDL, OCT2017) and a large real-world UF cohort, comparing against representative CNN and Transformer models.
Our experiments address three questions: 
(1) whether incorporating anatomical priors preserves classification performance; 
(2) whether explanations become more anatomically meaningful, assessed using anatomy-aware metrics and complementary faithfulness analyses; and 
(3) whether quantitative gains correspond to clinically interpretable behavior, examined through case studies and layer-wise attribution summaries. 
We further conduct ablations on feature fusion and attribution layer selection to isolate the effects of key design choices.

\subsection{Classification Performance}
We evaluate classification performance to verify that incorporating anatomical priors does not degrade disease detection accuracy. As summarized in Table~\ref{tab:auc_perf}, all models achieve near-saturated AUROC in public datasets, with many settings reaching $99$--$100\%$. In contrast, performance decreases in the real-world UF cohort (AUROC $89.6$--$98.1\%$), reflecting greater heterogeneity in EHR-derived cohorts.

Across tasks in UF cohort, RETFound yields the strongest overall performance, while SAIL models remain closely competitive. In particular, SAIL and SAIL-enc achieve AUROC within 0.3--0.5 points of RETFound across DME, AMD, Glaucoma, and ERM. These results confirm that the explanation improvements reported below are not driven by compromised predictive performance. The performance gap between public datasets and UF cohort further motivates evaluating explanation quality under realistic clinical conditions.

\begin{table}[htbp]
\centering
\caption{AUC performance (\%) of baseline and SAIL.}
\label{tab:auc_perf}
\small
\begin{tabular}{lcccccc}
\hline
\textbf{Dataset} & \textbf{RN} & \textbf{EN} & \textbf{ViT} & \textbf{RF} & \textbf{SAIL-enc} & \textbf{SAIL} \\
\hline
UF DME      & 96.8 & 95.5 & 96.1 & 98.1 & 97.7 & 97.8 \\
UF AMD      & 94.6 & 95.1 & 95.9 & 97.3 & 96.8 & 96.9 \\
UF Glaucoma & 89.6 & 90.1 & 92.2 & 94.1 & 93.6 & 93.7 \\
UF ERM      & 94.3 & 91.3 & 94.3 & 95.6 & 95.2 & 95.2 \\
OCTDL DME   & 99.5 & 97.6 & 99.6 & 100.0 & 100.0 & 100.0 \\
OCTDL AMD   & 99.7 & 99.0 & 99.1 & 99.6 & 99.7 & 99.8 \\
OCTDL ERM   & 99.7 & 94.3 & 100.0 & 99.3 & 99.9 & 99.3 \\
OCT2017 DME & 100.0 & 100.0 & 100.0 & 100.0 & 100.0 & 100.0 \\
\hline
\end{tabular}

\vspace{2pt}
\footnotesize
\textbf{Abbreviation:} RN: ResNet; EN: EfficientNet; RF: RETFound.
\end{table}

\subsection{Explainability Comparison}

\begin{figure*}[htbp]
    \centering
    \includegraphics[width=0.95\linewidth]{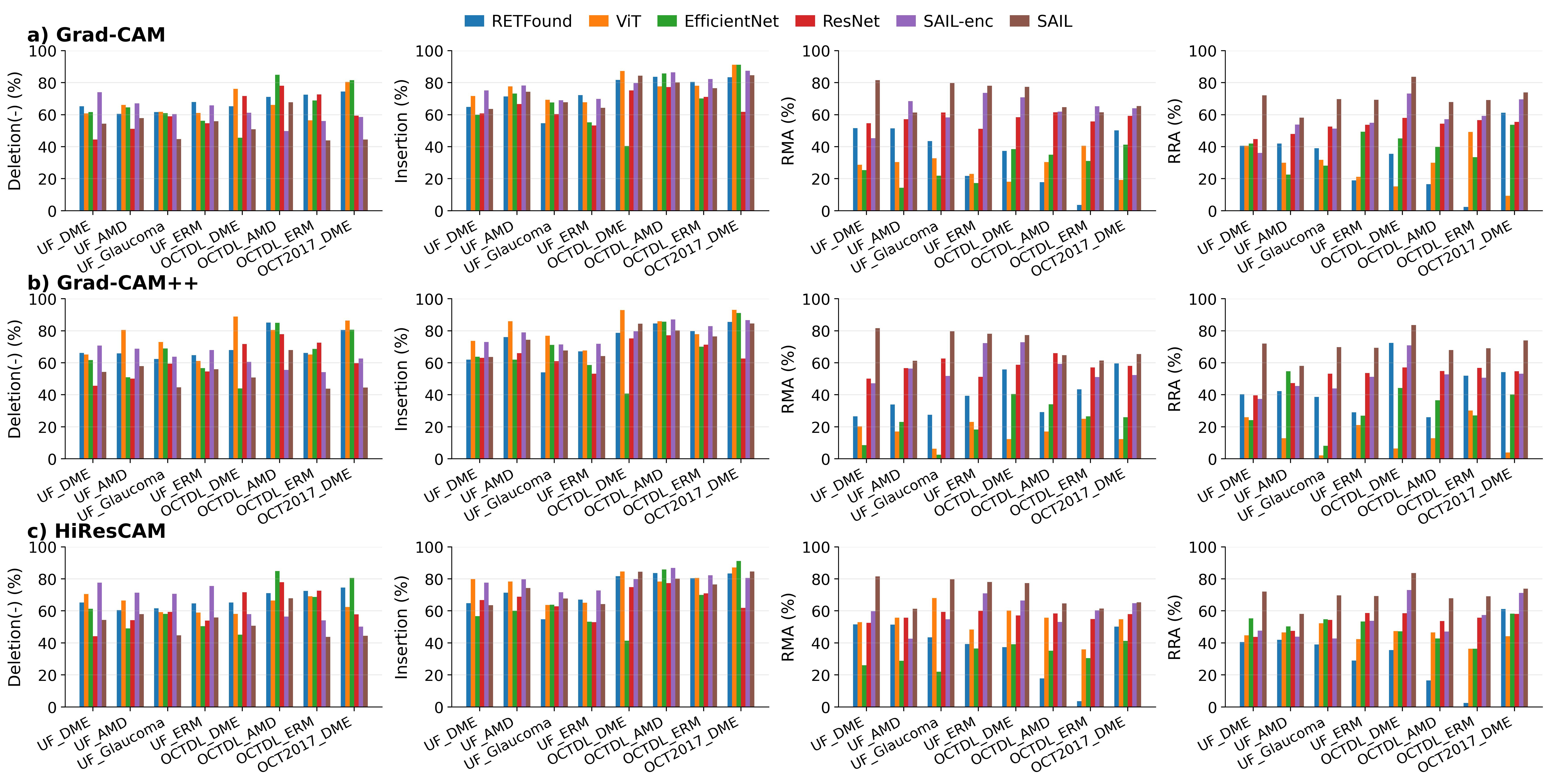}
    \caption{\textbf{Explainability performance for baseline and SAIL models.}
    (a) Using Grad-CAM. (b) Using Grad-CAM++. (c) Using HiResCAM.}
    \label{fig:rma_rra}
\end{figure*}

We test whether injecting anatomical priors into the representation (via segmentation pretraining and encoder–decoder fusion) improves where evidence is localized (RMA/RRA) while keeping faithfulness (Deletion/Insertion) competitive under the same post-hoc XAI methods.

Figure~\ref{fig:rma_rra} shows that SAIL consistently achieves the strongest anatomical alignment across datasets and XAI methods, with SAIL-enc typically ranking second. On UF DME under Grad-CAM (Figure~\ref{fig:rma_rra}-a), SAIL reaches 81.6/72.0 (RMA/RRA), exceeding the best baseline (ResNet) by more than 25 points on both metrics, indicating substantially more retina-confined attribution.

Similar gains are observed across other UF tasks and persist on public datasets. These trends remain stable under Grad-CAM++ (Figure~\ref{fig:rma_rra}-b) and HiResCAM (Figure~\ref{fig:rma_rra}-c), demonstrating robustness to XAI method choices.
In contrast, baseline models show sensitivity to the XAI methods. For example, RETFound and EfficientNet exhibit large alignment fluctuations across Grad-CAM variants, suggesting less stable localization behavior. ResNet is the most consistent baseline model but remains below SAIL models in anatomical alignment.

Faithfulness metrics show more mixed rankings and do not always correlate with RMA/RRA, reflecting their dependence on perturbation protocols and model-specific robustness. Nevertheless, SAIL and SAIL-enc remain competitive with the strongest baselines across deletion and insertion: Under Grad-CAM (Figure~\ref{fig:rma_rra}-a), SAIL often achieves best or second-best deletion while keeping comparable insertion, and SAIL-enc stays consistently strong across both metrics. Baselines show clearer specialization (e.g., ResNet and EfficientNet often stronger on deletion; RETFound and ViT often stronger on insertion), and these scores can be affected by model robustness.

Overall, SAIL delivers the clearest and most stable anatomical alignment gains, consistently concentrating attribution within retina. Meanwhile, SAIL-enc offers competitive faithfulness and anatomical alignment explanation.

\subsection{Qualitative Analysis}
\begin{figure}[!t]
    \centering
    \includegraphics[width=0.95\linewidth]{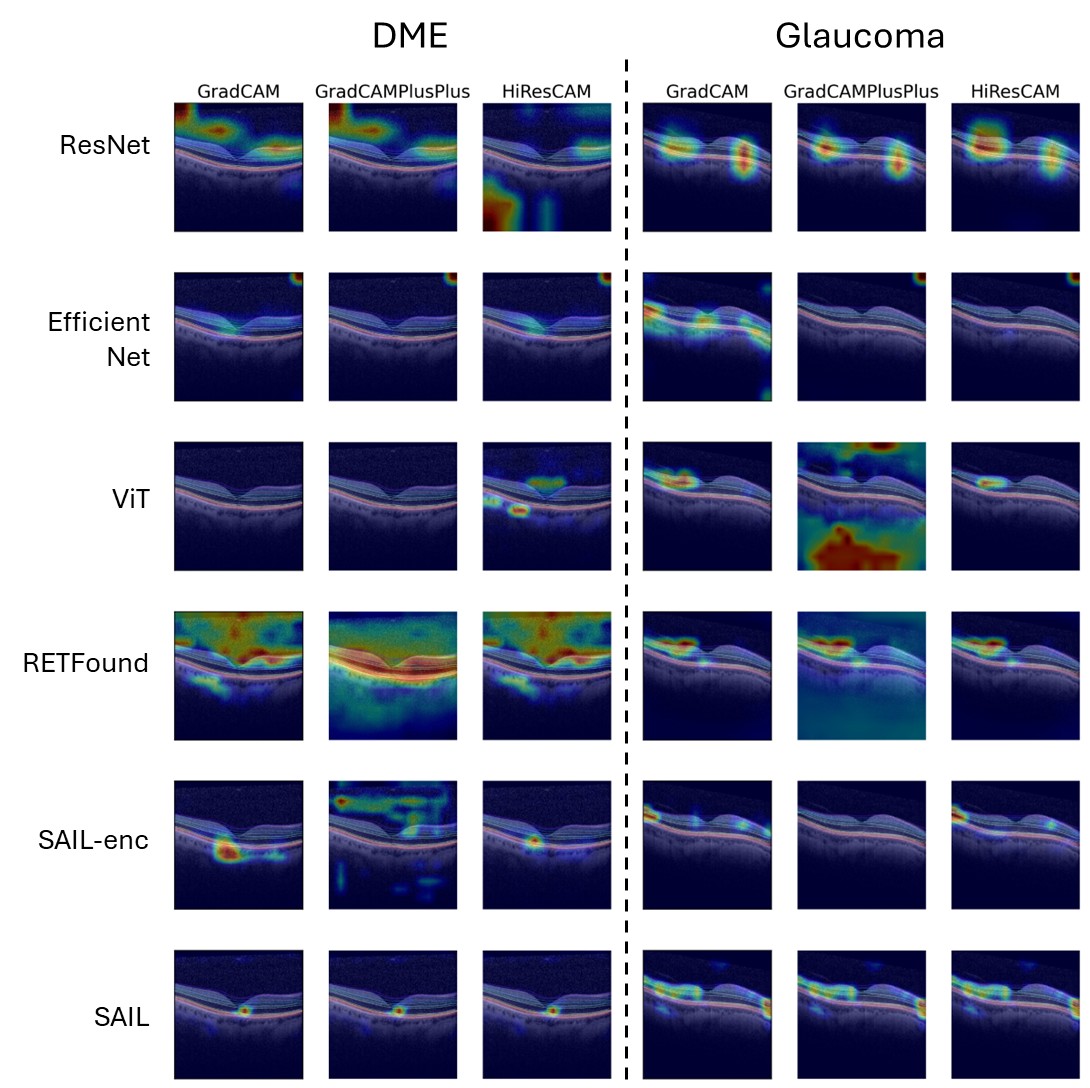}
    \caption{Case-study attribution map comparisons across diseases, models, and XAI methods.}
    \label{fig:casestudy}
\end{figure}

\subsubsection{Case Studies.}
Figure~\ref{fig:casestudy} presents representative B-scans for two tasks (DME and Glaucoma). Across diseases and Grad-CAM variants, SAIL produces the most anatomically grounded explanations: activations are largely confined to retinal tissue with minimal background or artifact leakage, consistent with its top RMA/RRA in Figure~\ref{fig:rma_rra}. In contrast, baselines exhibit broader and noisier attribution maps and background contamination, and are frequently dominated by non-diagnostic background signals.

For each disease, SAIL tends to form coherent, localized responses within the retina (often mid-to-lower layers for DME), while Glaucoma cases show more inner-to-mid retinal focus. Overall, these case studies support that structural priors improve specificity and anatomical validity. Full comparisons for all tasks are in Appendix~\ref{app:uf_process}.

\subsubsection{Layer Attribution Studies.}

\begin{table}[htbp]
\centering
\caption{SAIL layer attribution on the UF dataset.}
\label{tab:layer_attr_uf}
\begin{tabular}{l l cc}
\hline
\textbf{Task} & \textbf{Top-3 Layers} & \textbf{Top-3 Ratio} & \textbf{Top-3 Freq.} \\
\hline
DME      & ONL, IS/OS, OPL & 57.6\% & 51.2\% \\
AMD      & OSL, OPR, RPE   & 63.0\% & 71.6\% \\
Glaucoma & INL, OPL, ONL   & 58.7\% & 51.5\% \\
ERM      & GCL, IPL, RNFL  & 60.5\% & 60.1\% \\
\hline
\end{tabular}

\vspace{2pt}
\footnotesize
\textbf{Abbreviation:} Top-3 Freq.: Top-3 frequency; ONL: Outer Nuclear Layer; IS/OS: Inner Segment/Outer Segment; OPL: Outer Plexiform Layer; OSL: Outer Segment Layer; OPR: Outer Segment PR/PRE Complex; RPE: Retinal Pigment Epithelium; INL: Inner Nuclear Layer; GCL: Ganglion Cell Layer; IPL: Inner Plexiform Layer; RNFL: Nerve Fiber Layer.
\end{table}

To quantify where models attend, we pool attribution scores within OCTExplorer layer masks and report the Top-3 layers, Top-3 ratio (fraction of relevance mass in the top three layers), and Top-3 frequency (how often the same Top-3 set recurs).
As shown in Table 2, the Top-3 layers capture 58–63\% of the total relevance, indicating consistent concentration rather than uniform spread. AMD is the most concentrated and consistent (highest ratio and frequency) class, while DME and Glaucoma show more variability across patients. These layer-wise results complement the case studies: structure-aware explanations are not only retina-confined but also exhibit interpretable, task-dependent layer selectivity. 

\subsection{Ablation Studies}
We conduct ablation studies for (1) different feature source and fusion settings in SAIL and (2) which layer is explained, reporting RMA/RRA and Deletion/Insertion.

\subsubsection{Effect of the SAIL Feature Head.}
To diagnose the source of anatomy-aware explanation gains, we conduct ablation studies on feature source and fusion settings. Beyond the two main variants (SAIL and SAIL-enc), we additionally evaluate a decoder-only model (SAIL-dec) and several fusion operators, including channel merge, element-wise multiplication, and channel-wise modulation.

Table~\ref{tab:ablation_head} shows fusion is critical: SAIL achieves the best overall explainability with the highest alignment (RMA/RRA 81.6/72.0) while remaining faithfulness-competitive (Deletion/Insertion 54.3/63.6). SAIL-enc attains strong insertion but much weaker alignment, and SAIL-dec attains only modest alignment, suggesting a stronger influence from reconstruction-level cues. Among fusion operators, channel merge and channel-wise modulation provide moderate alignment gains, whereas multiplicative fusion is less stable and fails to yield meaningful improvements, often degrading insertion scores.

\begin{table}[htbp]
\centering
\caption{Ablation study of the classification head.}
\label{tab:ablation_head}
\setlength{\tabcolsep}{4pt}
\renewcommand{\arraystretch}{0.95}
\small
\begin{tabular}{lcccc}
\hline
\textbf{Model} & \textbf{Del. (-)} & \textbf{Insert} & \textbf{RMA} & \textbf{RRA} \\
\hline
SAIL-enc & 74.0\% & \textbf{75.1\%} & 45.3\% & 36.1\% \\
SAIL-dec & 58.2\% & 67.2\% & 50.3\% & 49.7\% \\
SAIL-merge & 58.7\% & 62.9\% & 76.9\% & 67.8\% \\
SAIL-multiply & 64.5\% & 58.1\% & 51.3\% & 56.3\% \\
SAIL-ch\_multiply & 67.1\% & 67.1\% & 70.6\% & 57.6\% \\
SAIL & \textbf{54.3\%} & 63.6\% & \textbf{81.6\%} & \textbf{72.0\%} \\
\hline
\end{tabular}

\vspace{2pt}
\footnotesize
\textbf{Abbreviation:} Del.: Deletion; Insert: Insertion; RMA: Relevance Mass Accuracy; RRA: Relevance Rank Accuracy; merge: channel merge; multiply: element-wise multiplication; ch\_multiply: channel-wise modulation.
\end{table}

\subsubsection{Effect of Feature Layer Selection.}
Table~\ref{tab:ablation_layers} shows a clear encoder-depth effect: explaining deeper encoder blocks improves alignment (e.g., SAIL-enc RMA increases from 25.0 to 45.3) while insertion remains relatively stable. For SAIL, attributing the fusion module provides the highest and most stable alignment (RMA 81.6), making it the most reliable target layer for explanation.

\begin{table}[htbp]
\centering
\caption{Ablation study of selecting different layers for XAI.}
\label{tab:ablation_layers}
\renewcommand{\arraystretch}{0.95}
\small
\begin{tabular}{llccc ccc}
\hline
\multirow{2}{*}{\textbf{Block}} & \multirow{2}{*}{\textbf{idx}} &
\multicolumn{3}{c}{\textbf{Insertion}} &
\multicolumn{3}{c}{\textbf{RMA}} \\
\cline{3-5}\cline{6-8}
 &  & \textbf{enc} & \textbf{dec} & \textbf{SAIL} & \textbf{enc} & \textbf{dec} & \textbf{SAIL} \\
\hline
\multirow{4}{*}{ENC}
& 0 & \textbf{77.9\%} & 77.9\% & 75.9\% & 25.0\% & 24.2\% & 24.2\% \\
& 1 & 77.0\% & 78.9\% & 74.6\% & 25.3\% & 22.7\% & 22.6\% \\
& 2 & 72.6\% & 76.9\% & 71.0\% & 44.3\% & 25.3\% & 62.8\% \\
& 3 & 75.1\% & \textbf{80.2\%} & 76.3\% & \textbf{45.3\%} & 34.6\% & 33.5\% \\
\hline
\multirow{5}{*}{DEC}
& 0 & -- & 71.8\% & 74.5\% & -- & 30.8\% & 24.6\% \\
& 1 & -- & 72.9\% & 72.0\% & -- & 55.7\% & 28.6\% \\
& 2 & -- & 72.0\% & \textbf{78.0\%} & -- & \textbf{59.5\%} & 17.4\% \\
& 3 & -- & 72.0\% & 76.4\% & -- & 55.3\% & 17.0\% \\
& 4 & -- & 71.5\% & 69.3\% & -- & 50.3\% & 11.8\% \\
\hline
Fusion
& 0 & -- & -- & 63.6\% & -- & -- & \textbf{81.6\%} \\
\hline
\end{tabular}

\vspace{2pt}
\footnotesize
\textbf{Abbreviation:} RMA: Relevance Mass Accuracy; ``--'': no selected layer;
enc: SAIL-enc; dec: SAIL-dec; ENC: encoder; DEC: decoder; Fusion: the fusion module.
\end{table}

\section{Discussion}
\label{sec:discussion}
We evaluate SAIL on public datasets and a large-scale real-world UF cohort. While predictive performance is near-ceiling on public datasets and degrades on UF due to clinical heterogeneity, models with similar accuracy exhibit markedly different explanation behaviors. Standard backbones often highlight non-diagnostic regions, whereas SAIL produces more anatomically grounded attribution.

SAIL consistently achieves the best anatomical alignment (RMA / RRA), with attribution concentrated within retinal tissue. Qualitative analysis confirms that these gains correspond to clearer evidence with reduced artifact and background leakage. Layer attribution complements pixel-level overlap by identifying which retinal layers drive decisions. On UF cohort, the Top-3 layers capture most relevance (>50\% Top-3 ratio). For DME, relevance concentrates in mid-retinal layers, including Outer Nuclear Layer (ONL), Inner Segment/Outer Segment (IS/OS), and Outer Plexiform Layer (OPL), which are frequently implicated in fluid-related disruption and structural staging \cite{dme1,dme2,dme3}. For AMD, relevance shifts toward RPE-associated regions, including Outer Segment Layer (OSL), Outer Segment PR/PRE Complex (OPR), and Retinal Pigment Epithelium (RPE), aligning with established OCT criteria used in AMD assessment \cite{amd1,amd2}. For Glaucoma, although Nerve Fiber Layer (RNFL) is a classic target \cite{glaucoma1,glaucoma2}, additional mid-layer relevance may reflect complementary inner retinal remodeling reported in prior studies \cite{glaucoma3,glaucoma4}. ERM shows dominance in inner layers, including Ganglion Cell Layer (GCL), Inner Plexiform Layer (IPL), and RNFL, consistent with the mechanical impact of epiretinal membranes on the inner retina \cite{erm1}. Ablations clarify that deeper encoder features yield more stable and better-aligned explanations, while decoder-only features produce flatter, less discriminative attribution. Fusion module is sensitive to operator choice, with some designs better preserving label-relevant semantics than others.

Our study has several limitations. First, our pipeline operates on single B-scans rather than full volumes, potentially missing off-center pathology. Second, RMA/RRA and layer attribution depend on automated segmentation, which can be noisy in severe disease or low-quality scans, potentially biasing alignment estimates. Third, RMA/RRA primarily reward attribution staying within retinal tissue and do not directly validate lesion or disease-specific correctness, and layer concentration reflects where evidence accumulates rather than clinical correctness. Future work will extend SAIL to volumetric modeling and conduct clinician co-evaluation to assess whether explanations translate into reliable decision support.

In sum, we propose SAIL, a structure-aware interpretable framework that integrates retinal anatomy into OCT model design. Under unchanged post-hoc XAI methods, SAIL produces more retina-aligned attribution, yielding explanations that are clearer and less artifact-driven. Layer-wise attribution and case studies provide clinically interpretable evidence by summarizing which retinal layers dominate model decisions. Retina-confined attribution and layer-localized summaries help clinicians quickly verify whether model attention aligns with known retinal biomarkers.

\section{Limitations and Ethical Considerations}
\label{sec:ethical}

This work uses retrospective UF Health data (IRB202300159) with a HIPAA Waiver of Authorization and safeguards to minimize privacy risk. A key limitation is geographic scope. The UF cohort is drawn from Florida, which may limit generalizability to other regions. The cohort is demographically diverse (22,036 patients; 59.78\% female; 60.88\% Non-Hispanic White, 24.61\% Non-Hispanic Black, 7.02\% Hispanic) and represent Florida population. SAIL is not a stand-alone diagnostic tool. Explanations can be over-trusted and should be used only with appropriate clinical oversight.

\begin{acks}
We would like to thank the University of Florida Research Computing (UFRC) for their support and the computational resources that made this research possible.
\end{acks}

\section{GenAI Disclosure}
Generative AI tools were used solely for language editing, including grammar correction, rephrasing for clarity, and terminology consistency checks.

\bibliographystyle{ACM-Reference-Format}
\bibliography{references}

@article{eye_image,
  title={The Eye as a Window to Systemic Health: A Survey of Retinal Imaging from Classical Techniques to Oculomics},
  author={Razzak, Imran and Jameel, Shoaib and others},
  journal={arXiv preprint arXiv:2505.04006},
  year={2025}
}

@article{oct_app,
  title={Retinal Thickness Analysis Using Optical Coherence Tomography: Diagnostic and Monitoring Applications in Retinal Diseases},
  author={Ahn, Seong Joon},
  journal={Diagnostics},
  volume={15},
  number={7},
  pages={833},
  year={2025},
  publisher={MDPI}
}

@article{oct_app2,
  title={Retinal optical coherence tomography imaging biomarkers: a review of the literature},
  author={Pandya, Bhadra U and Grinton, Michael and Mandelcorn, Efrem D and Felfeli, Tina},
  journal={Retina},
  volume={44},
  number={3},
  pages={369--380},
  year={2024},
  publisher={LWW}
}

@article{dl_overview,
  title={A comparison of deep learning performance against health-care professionals in detecting diseases from medical imaging: a systematic review and meta-analysis},
  author={Liu, Xiaoxuan and Faes, Livia and Kale, Aditya U and Wagner, Siegfried K and Fu, Dun Jack and Bruynseels, Alice and Mahendiran, Thushika and Moraes, Gabriella and Shamdas, Mohith and Kern, Christoph and others},
  journal={The lancet digital health},
  volume={1},
  number={6},
  pages={e271--e297},
  year={2019},
  publisher={Elsevier}
}

@article{dl_ophthalmology,
  title={Pivotal trial of an autonomous AI-based diagnostic system for detection of diabetic retinopathy in primary care offices},
  author={Abr{\`a}moff, Michael D and Lavin, Philip T and Birch, Michele and Shah, Nilay and Folk, James C},
  journal={NPJ digital medicine},
  volume={1},
  number={1},
  pages={39},
  year={2018},
  publisher={Nature Publishing Group UK London}
}

@article{dl_radiology,
  title={International evaluation of an AI system for breast cancer screening},
  author={McKinney, Scott Mayer and Sieniek, Marcin and Godbole, Varun and Godwin, Jonathan and Antropova, Natasha and Ashrafian, Hutan and Back, Trevor and Chesus, Mary and Corrado, Greg S and Darzi, Ara and others},
  journal={Nature},
  volume={577},
  number={7788},
  pages={89--94},
  year={2020},
  publisher={Nature Publishing Group UK London}
}

@article{XAI_problems,
  title={Benchmarking saliency methods for chest X-ray interpretation},
  author={Saporta, Adriel and Gui, Xiaotong and Agrawal, Ashwin and Pareek, Anuj and Truong, Steven QH and Nguyen, Chanh DT and Ngo, Van-Doan and Seekins, Jayne and Blankenberg, Francis G and Ng, Andrew Y and others},
  journal={Nature Machine Intelligence},
  volume={4},
  number={10},
  pages={867--878},
  year={2022},
  publisher={Nature Publishing Group UK London}
}

@article{XAI_problems_OCT,
  title={The role of saliency maps in enhancing ophthalmologists’ trust in artificial intelligence models},
  author={Wong, Carolyn Yu Tung and Antaki, Fares and Woodward-Court, Peter and Ong, Ariel Yuhan and Keane, Pearse A},
  journal={Asia-Pacific Journal of Ophthalmology},
  volume={13},
  number={4},
  pages={100087},
  year={2024},
  publisher={Elsevier}
}

@article{dl_XAI_review,
  title={Fostering trust and interpretability: integrating explainable AI (XAI) with machine learning for enhanced disease prediction and decision transparency},
  author={Agrawal, Renuka and Gupta, Tawishi and Gupta, Shaurya and Chauhan, Sakshi and Patel, Prisha and Hamdare, Safa},
  journal={Diagnostic Pathology},
  volume={20},
  number={1},
  pages={105},
  year={2025},
  publisher={Springer}
}

@article{lundberg2017unified,
  title={A unified approach to interpreting model predictions},
  author={Lundberg, Scott M and Lee, Su-In},
  journal={Advances in neural information processing systems},
  volume={30},
  year={2017}
}

@article{eye_image2,
  title={A feature transfer enabled multi-task deep learning model on medical imaging},
  author={Gao, Fei and Yoon, Hyunsoo and Wu, Teresa and Chu, Xianghua},
  journal={Expert Systems with Applications},
  volume={143},
  pages={112957},
  year={2020},
  publisher={Elsevier}
}

@article{eye_image3,
  title={Automated 3-D intraretinal layer segmentation of macular spectral-domain optical coherence tomography images},
  author={Garvin, Mona Kathryn and Abramoff, Michael David and Wu, Xiaodong and Russell, Stephen R and Burns, Trudy L and Sonka, Milan},
  journal={IEEE transactions on medical imaging},
  volume={28},
  number={9},
  pages={1436--1447},
  year={2009},
  publisher={IEEE}
}

@inproceedings{gradcam,
  title={Grad-cam: Visual explanations from deep networks via gradient-based localization},
  author={Selvaraju, Ramprasaath R and Cogswell, Michael and Das, Abhishek and Vedantam, Ramakrishna and Parikh, Devi and Batra, Dhruv},
  booktitle={Proceedings of the IEEE international conference on computer vision},
  pages={618--626},
  year={2017}
}

@inproceedings{gradcam++,
  title={Grad-cam++: Generalized gradient-based visual explanations for deep convolutional networks},
  author={Chattopadhay, Aditya and Sarkar, Anirban and Howlader, Prantik and Balasubramanian, Vineeth N},
  booktitle={2018 IEEE winter conference on applications of computer vision (WACV)},
  pages={839--847},
  year={2018},
  organization={IEEE}
}

@article{hirescam,
  title={Use HiResCAM instead of Grad-CAM for faithful explanations of convolutional neural networks},
  author={Draelos, Rachel Lea and Carin, Lawrence},
  journal={arXiv preprint arXiv:2011.08891},
  year={2020}
}

@inproceedings{lrp_2016,
  title={Layer-wise relevance propagation for neural networks with local renormalization layers},
  author={Binder, Alexander and Montavon, Gr{\'e}goire and Lapuschkin, Sebastian and M{\"u}ller, Klaus-Robert and Samek, Wojciech},
  booktitle={International conference on artificial neural networks},
  pages={63--71},
  year={2016},
  organization={Springer}
}

@article{crp,
  title={From attribution maps to human-understandable explanations through concept relevance propagation},
  author={Achtibat, Reduan and Dreyer, Maximilian and Eisenbraun, Ilona and Bosse, Sebastian and Wiegand, Thomas and Samek, Wojciech and Lapuschkin, Sebastian},
  journal={Nature Machine Intelligence},
  volume={5},
  number={9},
  pages={1006--1019},
  year={2023},
  publisher={Nature Publishing Group UK London}
}

@inproceedings{lime,
  author    = {Marco Tulio Ribeiro and
               Sameer Singh and
               Carlos Guestrin},
  title     = {"Why Should {I} Trust You?": Explaining the Predictions of Any Classifier},
  booktitle = {Proceedings of the 22nd {ACM} {SIGKDD} International Conference on
               Knowledge Discovery and Data Mining, San Francisco, CA, USA, August
               13-17, 2016},
  pages     = {1135--1144},
  year      = {2016},
}

@inproceedings{rise,
  title = {RISE: Randomized Input Sampling for Explanation of Black-box Models},
  author = {Vitali Petsiuk and Abir Das and Kate Saenko},
  booktitle = {Proceedings of the British Machine Vision Conference (BMVC)},
  year = {2018}
}

@INPROCEEDINGS{Transformer_Explainability,
  author={Chefer, Hila and Gur, Shir and Wolf, Lior},
  booktitle={2021 IEEE/CVF Conference on Computer Vision and Pattern Recognition (CVPR)}, 
  title={Transformer Interpretability Beyond Attention Visualization}, 
  year={2021},
  volume={},
  number={},
  pages={782-791},
  doi={10.1109/CVPR46437.2021.00084}
}

@article{OCT_XAI1,
  title={Diagnosis of multiple sclerosis using optical coherence tomography supported by explainable artificial intelligence},
  author={Dongil-Moreno, Francisco Javier and Ortiz, M and Pueyo, Ana and Boquete, L and S{\'a}nchez-Morla, Eva Mar{\'\i}a and Jimeno-Huete, Daniel and Miguel, JM and Barea, R and Vilades, Elisa and Garc{\'\i}a-Mart{\'\i}n, Elena},
  journal={Eye},
  volume={38},
  number={8},
  pages={1502--1508},
  year={2024},
  publisher={Nature Publishing Group UK London}
}

@article{OCT_XAI2,
  title={Explainable artificial intelligence toward usable and trustworthy computer-aided diagnosis of multiple sclerosis from Optical Coherence Tomography},
  author={Hernandez, Monica and Ramon-Julvez, Ubaldo and Vilades, Elisa and Cordon, Beatriz and Mayordomo, Elvira and Garcia-Martin, Elena},
  journal={PLoS One},
  volume={18},
  number={8},
  pages={e0289495},
  year={2023},
  publisher={Public Library of Science San Francisco, CA USA}
}

@article{OCT_XAI3,
  title={OCT-based diagnosis of glaucoma and glaucoma stages using explainable machine learning},
  author={Hasan, Md Mahmudul and Phu, Jack and Wang, Henrietta and Sowmya, Arcot and Kalloniatis, Michael and Meijering, Erik},
  journal={Scientific Reports},
  volume={15},
  number={1},
  pages={3592},
  year={2025},
  publisher={Nature Publishing Group UK London}
}

@article{OCT_reflect1,
  title={A clinically explainable AI-based grading system for age-related macular degeneration using optical coherence tomography},
  author={Elsharkawy, Mohamed and Sharafeldeen, Ahmed and Khalifa, Fahmi and Soliman, Ahmed and Elnakib, Ahmed and Ghazal, Mohammed and Sewelam, Ashraf and Thanos, Aristomenis and Sandhu, Harpal S and El-Baz, Ayman},
  journal={IEEE Journal of Biomedical and Health Informatics},
  volume={28},
  number={4},
  pages={2079--2090},
  year={2024},
  publisher={IEEE}
}

@inproceedings{oct_dl1,
  title={Interpretable retinal disease classification from oct images using deep neural network and explainable ai},
  author={Reza, Md Tanzim and Ahmed, Farzad and Sharar, Shihab and Rasel, Annajiat Alim},
  booktitle={2021 international conference on electronics, communications and information technology (ICECIT)},
  pages={1--4},
  year={2021},
  organization={IEEE}
}

@inproceedings{oct_dl2,
  title={Demystifying deep learning models for retinal OCT disease classification using explainable AI},
  author={Apon, Tasnim Sakib and Hasan, Mohammad Mahmudul and Islam, Abrar and Alam, Md Golam Rabiul},
  booktitle={2021 IEEE Asia-Pacific Conference on Computer Science and Data Engineering (CSDE)},
  pages={1--6},
  year={2021},
  organization={IEEE}
}

@article{oct_dlml,
  title={Development and validation of an explainable artificial intelligence framework for macular disease diagnosis based on optical coherence tomography images},
  author={Lv, Bin and Li, Shuang and Liu, Yang and Wang, Wei and Li, Hongyang and Zhang, Xiaoyue and Sha, Yanhui and Yang, Xiufen and Yang, Yang and Wang, Yue and others},
  journal={Retina},
  volume={42},
  number={3},
  pages={456--464},
  year={2022},
  publisher={LWW}
}

@article{retfound,
  title={A foundation model for generalizable disease detection from retinal images},
  author={Zhou, Yukun and Chia, Mark A and Wagner, Siegfried K and Ayhan, Murat S and Williamson, Dominic J and Struyven, Robbert R and Liu, Timing and Xu, Moucheng and Lozano, Mateo G and Woodward-Court, Peter and others},
  journal={Nature},
  volume={622},
  number={7981},
  pages={156--163},
  year={2023},
  publisher={Nature Publishing Group UK London}
}

@article{xai_metric,
  title={CLEVR-XAI: A benchmark dataset for the ground truth evaluation of neural network explanations},
  author={Arras, Leila and Osman, Ahmed and Samek, Wojciech},
  journal={Information Fusion},
  volume={81},
  pages={14--40},
  year={2022},
  publisher={Elsevier}
}

@inproceedings{xai_interpretable,
  title={Interpretable explanations of black boxes by meaningful perturbation},
  author={Fong, Ruth C and Vedaldi, Andrea},
  booktitle={Proceedings of the IEEE international conference on computer vision},
  pages={3429--3437},
  year={2017}
}

@article{octexplorer1,
  title={Retinal imaging and image analysis},
  author={Abr{\`a}moff, Michael D and Garvin, Mona K and Sonka, Milan},
  journal={IEEE reviews in biomedical engineering},
  volume={3},
  pages={169--208},
  year={2010},
  publisher={IEEE}
}

@article{octexplorer2,
  title={Optimal surface segmentation in volumetric images-a graph-theoretic approach},
  author={Li, Kang and Wu, Xiaodong and Chen, Danny Z and Sonka, Milan},
  journal={IEEE transactions on pattern analysis and machine intelligence},
  volume={28},
  number={1},
  pages={119--134},
  year={2006},
  publisher={IEEE}
}

@article{octexplorer3,
  title={Automated 3-D intraretinal layer segmentation of macular spectral-domain optical coherence tomography images},
  author={Garvin, Mona Kathryn and Abramoff, Michael David and Wu, Xiaodong and Russell, Stephen R and Burns, Trudy L and Sonka, Milan},
  journal={IEEE transactions on medical imaging},
  volume={28},
  number={9},
  pages={1436--1447},
  year={2009},
  publisher={IEEE}
}

@article{octdl,
  title={Octdl: Optical coherence tomography dataset for image-based deep learning methods},
  author={Kulyabin, Mikhail and Zhdanov, Aleksei and Nikiforova, Anastasia and Stepichev, Andrey and Kuznetsova, Anna and Ronkin, Mikhail and Borisov, Vasilii and Bogachev, Alexander and Korotkich, Sergey and Constable, Paul A and others},
  journal={Scientific data},
  volume={11},
  number={1},
  pages={365},
  year={2024},
  publisher={Nature Publishing Group UK London}
}

@article{oct2017,
  title={Identifying medical diagnoses and treatable diseases by image-based deep learning},
  author={Kermany, Daniel S and Goldbaum, Michael and Cai, Wenjia and Valentim, Carolina CS and Liang, Huiying and Baxter, Sally L and McKeown, Alex and Yang, Ge and Wu, Xiaokang and Yan, Fangbing and others},
  journal={cell},
  volume={172},
  number={5},
  pages={1122--1131},
  year={2018},
  publisher={Elsevier}
}

@article{dukedme,
  title={Kernel regression based segmentation of optical coherence tomography images with diabetic macular edema},
  author={Chiu, Stephanie J and Allingham, Michael J and Mettu, Priyatham S and Cousins, Scott W and Izatt, Joseph A and Farsiu, Sina},
  journal={Biomedical optics express},
  volume={6},
  number={4},
  pages={1172--1194},
  year={2015},
  publisher={Optical Society of America}
}

@article{nr206,
  title={Exploiting multi-granularity visual features for retinal layer segmentation in human eyes},
  author={He, Xiang and Wang, Yiming and Poiesi, Fabio and Song, Weiye and Xu, Quanqing and Feng, Zixuan and Wan, Yi},
  journal={Frontiers in Bioengineering and Biotechnology},
  volume={11},
  pages={1191803},
  year={2023},
  publisher={Frontiers Media SA}
}

@inproceedings{resnet,
  title={Deep residual learning for image recognition},
  author={He, Kaiming and Zhang, Xiangyu and Ren, Shaoqing and Sun, Jian},
  booktitle={Proceedings of the IEEE conference on computer vision and pattern recognition},
  pages={770--778},
  year={2016}
}

@article{vit,
  title={Visual transformers: Token-based image representation and processing for computer vision},
  author={Wu, Bichen and Xu, Chenfeng and Dai, Xiaoliang and Wan, Alvin and Zhang, Peizhao and Yan, Zhicheng and Tomizuka, Masayoshi and Gonzalez, Joseph and Keutzer, Kurt and Vajda, Peter},
  journal={arXiv preprint arXiv:2006.03677},
  year={2020}
}

@inproceedings{effnet,
  title={Efficientnet: Rethinking model scaling for convolutional neural networks},
  author={Tan, Mingxing and Le, Quoc},
  booktitle={International conference on machine learning},
  pages={6105--6114},
  year={2019},
  organization={PMLR}
}

@article{dme1,
  title={Diabetic macular oedema—need for a unified consensus classification based on clinical and imaging features},
  author={Kumawat, Devesh and Venkatesh, Pradeep},
  journal={Eye Open},
  volume={2},
  number={1},
  pages={2},
  year={2026},
  publisher={Nature Publishing Group UK London}
}

@article{dme2,
  title={Structural changes in individual retinal layers in diabetic macular edema},
  author={Murakami, Tomoaki and Yoshimura, Nagahisa},
  journal={Journal of diabetes research},
  volume={2013},
  number={1},
  pages={920713},
  year={2013},
  publisher={Wiley Online Library}
}

@article{dme3,
  title={Explainable Artificial Intelligence-Assisted Exploration of Clinically Significant Diabetic Retinal Neurodegeneration on OCT Images},
  author={Yoshida, Miyo and Murakami, Tomoaki and Ishihara, Kenji and Mori, Yuki and Tsujikawa, Akitaka},
  journal={Ophthalmology Science},
  pages={100804},
  year={2025},
  publisher={Elsevier}
}

@article{amd1,
  title={Consensus definition for atrophy associated with age-related macular degeneration on OCT: classification of atrophy report 3},
  author={Sadda, Srinivas R and Guymer, Robyn and Holz, Frank G and Schmitz-Valckenberg, Steffen and Curcio, Christine A and Bird, Alan C and Blodi, Barbara A and Bottoni, Ferdinando and Chakravarthy, Usha and Chew, Emily Y and others},
  journal={Ophthalmology},
  volume={125},
  number={4},
  pages={537--548},
  year={2018},
  publisher={Elsevier}
}

@article{amd2,
  title={Age-Related Macular Degeneration Preferred Practice Pattern{\textregistered}},
  author={Vemulakonda, G Atma and Bailey, Steven T and Kim, Stephen J and Kovach, Jaclyn L and Lim, Jennifer I and Ying, Gui-shuang and Flaxel, Christina J and others},
  journal={Ophthalmology},
  volume={132},
  number={4},
  pages={P1--P74},
  year={2025}
}

@article{glaucoma1,
  title={OCT Glaucoma Staging System: a new method for retinal nerve fiber layer damage classification using spectral-domain OCT},
  author={Brusini, P},
  journal={Eye},
  volume={32},
  number={1},
  pages={113--119},
  year={2018},
  publisher={Nature Publishing Group}
}

@article{glaucoma2,
  title={Primary open-angle glaucoma preferred practice pattern{\textregistered}},
  author={Gedde, Steven J and Vinod, Kateki and Wright, Martha M and Muir, Kelly W and Lind, John T and Chen, Philip P and Li, Tianjing and Mansberger, Steven L},
  journal={Ophthalmology},
  volume={128},
  number={1},
  pages={P71--P150},
  year={2021},
  publisher={Elsevier}
}

@article{glaucoma3,
  title={Thicker Inner Nuclear Layer as a Predictor of Glaucoma Progression and the Impact of Intraocular Pressure Fluctuation},
  author={Jung, Kyoung In and Ryu, Hee Kyung and Oh, Si Eun and Shin, Hee Jong and Park, Chan Kee},
  journal={Journal of Clinical Medicine},
  volume={13},
  number={8},
  pages={2312},
  year={2024},
  publisher={MDPI}
}

@article{glaucoma4,
  title={Transverse separation of the outer retinal layer at the peripapillary in glaucomatous myopes},
  author={Kim, Yong Chan and Hwang, Ho Sik and Park, Hae-Young Lopilly and Park, Chan Kee},
  journal={Scientific Reports},
  volume={8},
  number={1},
  pages={12446},
  year={2018},
  publisher={Nature Publishing Group UK London}
}

@article{erm1,
  title={Interpretable detection of epiretinal membrane from optical coherence tomography with deep neural networks},
  author={Ayhan, Murat Se{\c{c}}kin and Neubauer, Jonas and Uzel, Mehmet Murat and Gelisken, Faik and Berens, Philipp},
  journal={Scientific Reports},
  volume={14},
  number={1},
  pages={8484},
  year={2024},
  publisher={Nature Publishing Group UK London}
}

@article{burton2021lancet,
  title={The lancet global health commission on global eye health: vision beyond 2020},
  author={Burton, Matthew J and Ramke, Jacqueline and Marques, Ana Patricia and Bourne, Rupert RA and Congdon, Nathan and Jones, Iain and Tong, Brandon AM Ah and Arunga, Simon and Bachani, Damodar and Bascaran, Covadonga and others},
  journal={The Lancet Global Health},
  volume={9},
  number={4},
  pages={e489--e551},
  year={2021},
  publisher={Elsevier}
}

@article{li2023global,
  title={The global incidence and disability of eye injury: an analysis from the Global Burden of Disease Study 2019},
  author={Li, Cong and Fu, Yongyan and Liu, Shunming and Yu, Honghua and Yang, Xiaohong and Zhang, Meixia and Liu, Lei},
  journal={EClinicalMedicine},
  volume={62},
  year={2023},
  publisher={Elsevier}
}

@article{zhang2024global,
  title={Global burden of low vision and blindness due to age-related macular degeneration from 1990 to 2021 and projections for 2050},
  author={Zhang, Shiyan and Ren, Jianping and Chai, Ruiting and Yuan, Shuang and Hao, Yinzhu},
  journal={BMC Public Health},
  volume={24},
  number={1},
  pages={3510},
  year={2024},
  publisher={Springer}
}

@article{zhou2023visual,
  title={Visual impairment and blindness caused by retinal diseases: A nationwide register-based study},
  author={Zhou, Chuandi and Li, Shu and Ye, Luyao and Chen, Chong and Liu, Shu and Yang, Hongxia and Zhuang, Peng and Liu, Zengye and Jiang, Hongwen and Han, Jing and others},
  journal={Journal of Global Health},
  volume={13},
  pages={04126},
  year={2023}
}

@article{lim2025diabetic,
  title={Diabetic Retinopathy Preferred Practice Pattern{\textregistered}},
  author={Lim, Jennifer I and Kim, Stephen J and Bailey, Steven T and Kovach, Jaclyn L and Vemulakonda, G Atma and Ying, Gui-shuang and Flaxel, Christina J and others},
  journal={Ophthalmology},
  volume={132},
  number={4},
  pages={P75--P162},
  year={2025}
}

@article{wilkinson2003proposed,
  title={Proposed international clinical diabetic retinopathy and diabetic macular edema disease severity scales},
  author={Wilkinson, Charles P and Ferris III, Frederick L and Klein, Ronald E and Lee, Paul P and Agardh, Carl David and Davis, Matthew and Dills, Diana and Kampik, Anselm and Pararajasegaram, Rangasamy and Verdaguer, Juan T and others},
  journal={Ophthalmology},
  volume={110},
  number={9},
  pages={1677--1682},
  year={2003},
  publisher={Elsevier}
}

@article{de2018clinically,
  title={Clinically applicable deep learning for diagnosis and referral in retinal disease},
  author={De Fauw, Jeffrey and Ledsam, Joseph R and Romera-Paredes, Bernardino and Nikolov, Stanislav and Tomasev, Nenad and Blackwell, Sam and Askham, Harry and Glorot, Xavier and O’Donoghue, Brendan and Visentin, Daniel and others},
  journal={Nature medicine},
  volume={24},
  number={9},
  pages={1342--1350},
  year={2018},
  publisher={Nature Publishing Group US New York}
}

@article{grace2021investigation,
  title={Investigation of the efficacy of an online tool for improving the diagnosis of macular lesions imaged by optical coherence tomography},
  author={Grace, Paul and Evans, Bruce JW and Edgar, David F and Patel, Praveen J and Thomas, Dhanes and Mahon, Gerald and Blake, Alison and Bennett, David},
  journal={Journal of Optometry},
  volume={14},
  number={2},
  pages={206--214},
  year={2021},
  publisher={Elsevier}
}

@article{markan2020novel,
  title={Novel imaging biomarkers in diabetic retinopathy and diabetic macular edema},
  author={Markan, Ashish and Agarwal, Aniruddha and Arora, Atul and Bazgain, Krinjeela and Rana, Vipin and Gupta, Vishali},
  journal={Therapeutic Advances in Ophthalmology},
  volume={12},
  pages={2515841420950513},
  year={2020},
  publisher={SAGE Publications Sage UK: London, England}
}

@article{xai_comp,
  title={A Comprehensive Review of Explainable Artificial Intelligence (XAI) in Computer Vision},
  author={Cheng, Zhihan and Wu, Yue and Li, Yule and Cai, Lingfeng and Ihnaini, Baha},
  journal={Sensors},
  volume={25},
  number={13},
  pages={4166},
  year={2025},
  publisher={MDPI}
}

@article{gao2020feature,
  title={A feature transfer enabled multi-task deep learning model on medical imaging},
  author={Gao, Fei and Yoon, Hyunsoo and Wu, Teresa and Chu, Xianghua},
  journal={Expert Systems with Applications},
  volume={143},
  pages={112957},
  year={2020},
  publisher={Elsevier}
}

@article{zhao2023multi,
  title={Multi-task deep learning for medical image computing and analysis: A review},
  author={Zhao, Yan and Wang, Xiuying and Che, Tongtong and Bao, Guoqing and Li, Shuyu},
  journal={Computers in Biology and Medicine},
  volume={153},
  pages={106496},
  year={2023},
  publisher={Elsevier}
}

@article{shi2024retinal,
  title={Retinal structure guidance-and-adaption network for early Parkinson’s disease recognition based on OCT images},
  author={Shi, Hanfeng and Wei, Jiaqi and Jin, Richu and Peng, Jiaxin and Wang, Xingyue and Hu, Yan and Zhang, Xiaoqing and Liu, Jiang},
  journal={Computerized Medical Imaging and Graphics},
  volume={118},
  pages={102463},
  year={2024},
  publisher={Elsevier}
}

@article{wen2024concept,
  title={Concept-based lesion aware transformer for interpretable retinal disease diagnosis},
  author={Wen, Chi and Ye, Mang and Li, He and Chen, Ting and Xiao, Xuan},
  journal={IEEE Transactions on Medical Imaging},
  year={2024},
  publisher={IEEE}
}

@article{zhang2025retinal,
  title={Retinal OCT image segmentation with deep learning: A review of advances, datasets, and evaluation metrics},
  author={Zhang, Huihong and Yang, Bing and Li, Sanqian and Zhang, Xiaoqing and Li, Xiaoling and Liu, Tianhang and Higashita, Risa and Liu, Jiang},
  journal={Computerized Medical Imaging and Graphics},
  pages={102539},
  year={2025},
  publisher={Elsevier}
}

@inproceedings{long2015fully,
  title={Fully convolutional networks for semantic segmentation},
  author={Long, Jonathan and Shelhamer, Evan and Darrell, Trevor},
  booktitle={Proceedings of the IEEE conference on computer vision and pattern recognition},
  pages={3431--3440},
  year={2015}
}

@inproceedings{unet,
  title={U-net: Convolutional networks for biomedical image segmentation},
  author={Ronneberger, Olaf and Fischer, Philipp and Brox, Thomas},
  booktitle={International Conference on Medical image computing and computer-assisted intervention},
  pages={234--241},
  year={2015},
  organization={Springer}
}

@article{fang2017automatic,
  title={Automatic segmentation of nine retinal layer boundaries in OCT images of non-exudative AMD patients using deep learning and graph search},
  author={Fang, Leyuan and Cunefare, David and Wang, Chong and Guymer, Robyn H and Li, Shutao and Farsiu, Sina},
  journal={Biomedical optics express},
  volume={8},
  number={5},
  pages={2732--2744},
  year={2017},
  publisher={Optical Society of America}
}

@article{relaynet,
  title={ReLayNet: retinal layer and fluid segmentation of macular optical coherence tomography using fully convolutional networks},
  author={Roy, Abhijit Guha and Conjeti, Sailesh and Karri, Sri Phani Krishna and Sheet, Debdoot and Katouzian, Amin and Wachinger, Christian and Navab, Nassir},
  journal={Biomedical optics express},
  volume={8},
  number={8},
  pages={3627--3642},
  year={2017},
  publisher={Optical Society of America}
}

@article{cenet,
  title={Ce-net: Context encoder network for 2d medical image segmentation},
  author={Gu, Zaiwang and Cheng, Jun and Fu, Huazhu and Zhou, Kang and Hao, Huaying and Zhao, Yitian and Zhang, Tianyang and Gao, Shenghua and Liu, Jiang},
  journal={IEEE transactions on medical imaging},
  volume={38},
  number={10},
  pages={2281--2292},
  year={2019},
  publisher={IEEE}
}

@article{tan2023retinal,
  title={Retinal layer segmentation in OCT images with boundary regression and feature polarization},
  author={Tan, Yubo and Shen, Wen-Da and Wu, Ming-Yuan and Liu, Gui-Na and Zhao, Shi-Xuan and Chen, Yang and Yang, Kai-Fu and Li, Yong-Jie},
  journal={IEEE Transactions on Medical Imaging},
  volume={43},
  number={2},
  pages={686--700},
  year={2023},
  publisher={IEEE}
}

@article{chen2021transunet,
  title={Transunet: Transformers make strong encoders for medical image segmentation},
  author={Chen, Jieneng and Lu, Yongyi and Yu, Qihang and Luo, Xiangde and Adeli, Ehsan and Wang, Yan and Lu, Le and Yuille, Alan L and Zhou, Yuyin},
  journal={arXiv preprint arXiv:2102.04306},
  year={2021}
}

@inproceedings{zhong2025unioctseg,
  title={UniOCTSeg: Towards Universal OCT Retinal Layer Segmentation via Hierarchical Prompting and Progressive Consistency Learning},
  author={Zhong, Jian and Lin, Li and Miao, Chaoran and Wong, Kenneth KY and Tang, Xiaoying},
  booktitle={International Conference on Medical Image Computing and Computer-Assisted Intervention},
  pages={629--639},
  year={2025},
  organization={Springer}
}

@article{wang2025structural,
  title={Structural-prior guided and feature-enhanced transformer with masked image modeling pretraining for retinal layers and fluid segmentation in macular edema OCT images},
  author={Wang, Sheng and Feng, Shuxian and Wang, Zhina and Ji, Zhenning and Liu, Jiajia and Chen, Wei and Fu, Binzhe and Liu, Rong and Chen, Wenliang and Dai, Yining and others},
  journal={Biomedical Optics Express},
  volume={16},
  number={12},
  pages={5096--5117},
  year={2025},
  publisher={Optica Publishing Group}
}

@article{jain2019attention,
  title={Attention is not explanation},
  author={Jain, Sarthak and Wallace, Byron C},
  journal={arXiv preprint arXiv:1902.10186},
  year={2019}
}

@inproceedings{chefer2021transformer,
  title={Transformer interpretability beyond attention visualization},
  author={Chefer, Hila and Gur, Shir and Wolf, Lior},
  booktitle={Proceedings of the IEEE/CVF conference on computer vision and pattern recognition},
  pages={782--791},
  year={2021}
}

@article{abnar2020,
  author       = {Samira Abnar and
                  Willem H. Zuidema},
  title        = {Quantifying Attention Flow in Transformers},
  journal      = {CoRR},
  volume       = {abs/2005.00928},
  year         = {2020},
  url          = {https://arxiv.org/abs/2005.00928},
  eprinttype    = {arXiv},
  eprint       = {2005.00928},
  bibsource    = {dblp computer science bibliography, https://dblp.org}
}

@inproceedings{hooper2023,
 author = {Hooper, Sarah and Chen, Mayee and Saab, Khaled and Bhatia, Kush and Langlotz, Curtis and R\'{e}, Christopher},
 booktitle = {Advances in Neural Information Processing Systems},
 editor = {A. Oh and T. Naumann and A. Globerson and K. Saenko and M. Hardt and S. Levine},
 pages = {55415--55441},
 publisher = {Curran Associates, Inc.},
 title = {A case for reframing automated medical image classification as segmentation},
 url = {https://proceedings.neurips.cc/paper_files/paper/2023/file/ad6a3bd12095fdca71c306871bdec400-Paper-Conference.pdf},
 volume = {36},
 year = {2023}
}

@misc{Iakubovskii2019,
  Author = {Pavel Iakubovskii},
  Title = {Segmentation Models Pytorch},
  Year = {2019},
  Publisher = {GitHub},
  Journal = {GitHub repository},
  Howpublished = {\url{https://github.com/qubvel/segmentation_models.pytorch}}
}

\appendix

\section{UF Dataset Processing \& Statistics}
\label{app:uf_process}
\subsection{UF Dataset Processing}

\begin{figure}[htbp]
    \centering
    \includegraphics[width=\linewidth]{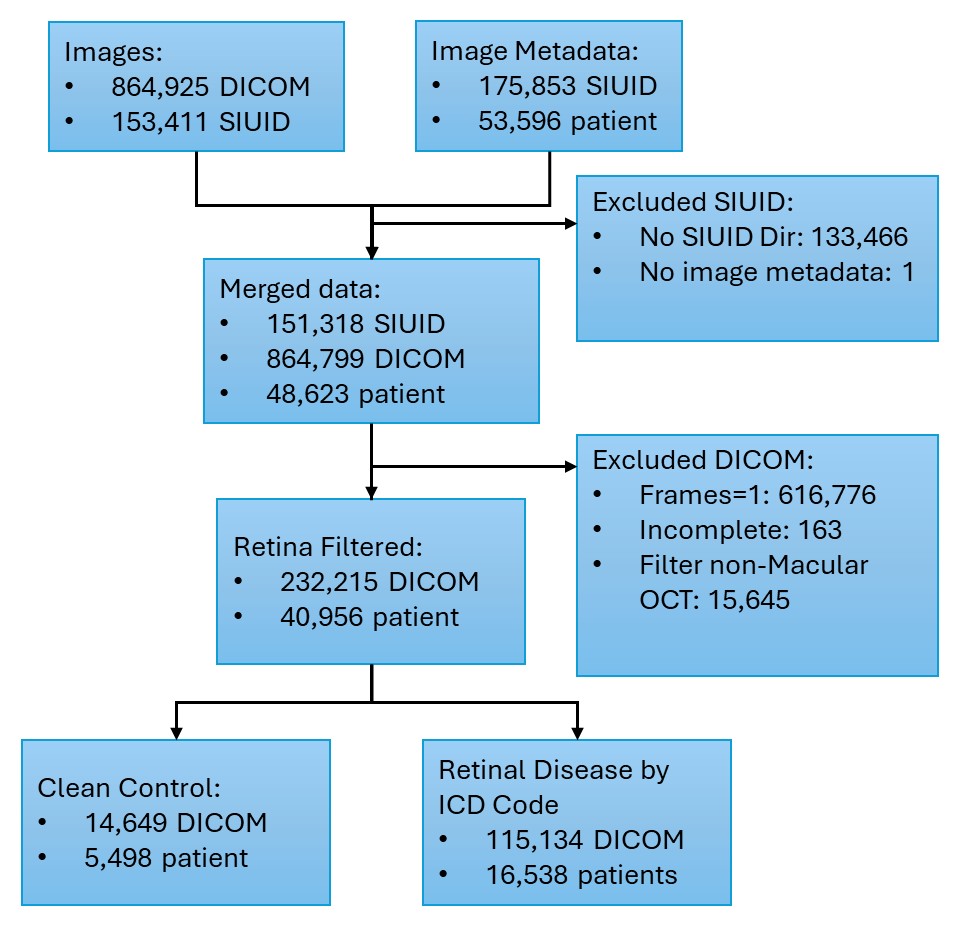}
    \caption{Data curation and cohort construction pipeline for retinal OCT.}
    \label{fig:dataprocess}
\end{figure}

Figure~\ref{fig:dataprocess} illustrates the full data curation and cohort construction workflow for the UF real-world OCT dataset. The source data are derived from the UF Health clinical data repository, where structured electronic health record (EHR) information has been harmonized into the OMOP Common Data Model (CDM). For this study, we focus on diagnosis-linked retinal OCT imaging, pairing each imaging study with relevant clinical identifiers for downstream model development and evaluation. All records are de-identified using randomly generated unique identifiers to ensure privacy and data security.

The raw imaging archive contains 864,925 DICOM files corresponding to 153,411 Study Instance UIDs (SIUIDs), along with associated metadata covering 175,853 SIUIDs from 53,596 patients. After merging the imaging data and metadata by SIUID, we obtained 151,318 SIUIDs comprising 864,799 DICOM files from 48,623 patients. SIUIDs without a corresponding study directory (133,466 SIUIDs) or missing metadata (1 SIUID) are excluded. We then filter DICOM files by validity and modality, removing studies with only one frame (616,776 DICOMs), incomplete entries (163 DICOMs), and non-macular OCT scans (15,645 DICOMs). Later, we derive classification labels from the patient’s diagnosis history using ICD-9/ICD-10 codes recorded in the EHR. For control patients, we select the clean control patients who don’t have retinal-related disease and glaucoma ICD codes. The clean control dataset contains 14,649 DICOM from 5,498 patients, and the retinal disease dataset contains 115,134 total DICOM from 16,538 total patients.

For supervised learning, we derive classification labels from the patient’s diagnosis history using ICD-9/ICD-10 codes recorded in EHR. For each disease category, we define a rule-based mapping from ICD codes and laterality (left, right, or both eyes) to labels, and drop the latency not included to avoid labeling errors. For OCT B-scans from control patients, we select the clean control patients who do not have retinal-related disease or glaucoma ICD codes. The detailed ICD codes for disease and laterality are in Table~\ref{tab:icd_laterality}.

\begin{table}[htbp]
\centering
\caption{ICD codes used for task definitions by laterality.}
\label{tab:icd_laterality}
\setlength{\tabcolsep}{4pt}
\renewcommand{\arraystretch}{1.05}
\footnotesize
\begin{tabular}{l l p{0.68\linewidth}}
\hline
\textbf{Task} & \textbf{Laterality} & \textbf{ICD Used} \\
\hline
\multirow{3}{*}{AMD}
& both  & H35.313, H35.3130, H35.3131, H35.3132, H35.3133, H35.3134, H35.32, H35.3230, H35.3231, H35.3232, H35.3233, 362.52 \\
& left  & H35.312, H35.3120, H35.3121, H35.3122, H35.3123, H35.3124, H35.322, H35.3220, H35.3221, H35.3222, H35.3223 \\
& right & H35.311, H35.3110, H35.3111, H35.3112, H35.3113, H35.3114, H35.321, H35.3210, H35.3211, H35.3212, H35.3213 \\
\hline
\multirow{3}{*}{DME}
& both  & E10.3213, E11.3213, E10.3313, E11.3313, E10.3413, E11.3413, E10.3513, E11.3513, 362.07 \\
& left  & E10.3212, E11.3212, E10.3312, E11.3312, E10.3412, E11.3412, E10.3512, E11.3512 \\
& right & E10.3211, E11.3211, E10.3311, E11.3311, E10.3411, E11.3411, E10.3511, E11.3511 \\
\hline
\multirow{3}{*}{ERM}
& both  & H35.373, 362.56 \\
& left  & H35.372 \\
& right & H35.371 \\
\hline
\multirow{3}{*}{Glaucoma}
& both  & H40.1134, H40.1131, H40.1132, H40.1133, 365.11 \\
& left  & H40.1124, H40.1121, H40.1122, H40.1123 \\
& right & H40.1114, H40.1111, H40.1112, H40.1113 \\
\hline
\end{tabular}
\end{table}

\subsection{UF Dataset Statistics}

As shown in Table~\ref{tab:demographics}, the UF real-world cohort contains 115,134 OCT DICOMs from 22,036 unique patients. The cohort is 59.78\% female (13,174) and 40.22\% male (8,862). In terms of race/ethnicity, patients are predominantly non-Hispanic White (60.88\%, 13,416), followed by non-Hispanic Black (24.61\%, 5,423) and Hispanic (7.02\%, 1,548), with Other (5.95\%, 1,312) and Unknown (1.53\%, 337) categories. The mean age is 62.31 years (SD 16.38), with a median of 65 years.

\begin{table}[htbp]
\centering
\caption{Cohort demographic distribution.}
\label{tab:demographics}
\setlength{\tabcolsep}{5pt}
\renewcommand{\arraystretch}{0.95}
\small
\begin{tabular}{llcc}
\hline
\textbf{Demographic} & \textbf{Group} & \textbf{Patients} & \textbf{Percent} \\
\hline
\multirow{2}{*}{Gender}
& FEMALE & 13{,}174 & 59.78\% \\
& MALE   & 8{,}862  & 40.22\% \\
\hline
\multirow{5}{*}{Race/Ethnicity}
& HISPANIC & 1{,}548  & 7.02\% \\
& NHB      & 5{,}423  & 24.61\% \\
& NHW      & 13{,}416 & 60.88\% \\
& OTHER    & 1{,}312  & 5.95\% \\
& UNKNOWN  & 337      & 1.53\% \\
\hline
\end{tabular}
\end{table}

\section{Segmentation Pretraining}
\label{app:segmentation}

\subsection{Related Work for OCT Segmentation}
Deep learning–based image segmentation provides an effective mechanism for learning spatially grounded representations in domains where downstream tasks depend on anatomically meaningful structures. In retinal optical coherence tomography (OCT) imaging, segmentation is particularly well suited for this purpose due to the highly organized laminar structure of the retina and choroid. Retinal layer segmentation offers explicit structural supervision and has been widely leveraged for disease analysis, progression assessment, and biomarker discovery \cite{zhang2025retinal}.

Most retinal OCT segmentation methods are based on convolutional neural networks (CNNs) with encoder–decoder architectures. Early work adapted Fully Convolutional Networks (FCNs) to OCT images, demonstrating the feasibility of end-to-end retinal layer segmentation without handcrafted features \cite{long2015fully}. Subsequently, U-Net and its variants became the dominant architectural paradigm due to their ability to preserve fine anatomical boundaries through multi-scale feature fusion enabled by skip connections \cite{unet}. These CNN-based designs have consistently demonstrated strong performance across both normal and pathological OCT scans, including diabetic macular edema and age-related macular degeneration, even in the presence of noise, shadowing artifacts, and ambiguous layer boundaries \cite{fang2017automatic,relaynet}. Beyond standard U-shaped architectures, many OCT-specific models introduce moderate architectural refinements while retaining a convolutional backbone. These extensions aim to improve boundary localization, incorporate multi-scale contextual information, or enhance robustness to pathological deformation, while preserving the strong spatial inductive biases inherent to convolutional operators \cite{cenet,tan2023retinal}. As a result, CNN-based architecture remains the most widely adopted and reproducible baselines across retinal OCT segmentation benchmarks and comparative studies \cite{zhang2025retinal}.

More recently, Transformer-based and hybrid CNN–Transformer architectures have been explored for retinal OCT segmentation, motivated by the ability of self-attention mechanisms to capture long-range dependencies. Recent studies integrate Transformer modules into convolutional encoders to enhance global context modeling while maintaining local structural sensitivity and have reported promising results on OCT layer and pathology segmentation tasks \cite{chen2021transunet,zhong2025unioctseg,wang2025structural}. Despite encouraging segmentation accuracy, Transformer-style representations are less directly aligned with anatomy-consistent, pixel-level attribution. Attention weights are distributed across layers and heads and do not necessarily correspond to feature importance, limiting their suitability for anatomy-aware interpretation using widely adopted gradient-based explainability methods \cite{jain2019attention,chefer2021transformer}. While recent work proposes attention aggregation and propagation schemes, such as attention rollout or attention flow, to better approximate token-level relevance \cite{abnar2020}, their anatomical faithfulness and stability in pixel-level medical imaging settings have not been systematically validated and remain an open practical concern.

From a representation learning perspective, convolutional encoders impose a locality-preserving inductive bias that encourages learned features to remain spatially grounded throughout the network hierarchy. This property is particularly important for downstream interpretation, as it preserves a direct correspondence between intermediate feature activations and anatomical structures. In contrast, token-based representations in Transformer architectures introduce an additional abstraction layer between image space and decision space, which may obscure spatial correspondence when explanations are projected back to pixel space. Given these considerations, we adopt a CNN-based segmentation framework as the foundation for anatomical structure learning in OCT. This choice aligns with dominant practice in retinal OCT segmentation while explicitly constraining the learned representations to preserve spatial correspondence between features and retinal anatomy. Such spatially grounded representations are particularly well suited for downstream classification explainability, enabling attribution maps that can be interpreted in a causally plausible and anatomically meaningful manner. More broadly, this design choice is consistent with recent perspectives advocating segmentation-based or structure-supervised learning as a principled foundation for medical image classification, particularly when anatomical localization is clinically meaningful \cite{hooper2023}. By pretraining on segmentation tasks, the encoder is biased toward structure-aware representations that transfer to downstream classification and interpretation without introducing additional ambiguity associated with attention-based attribution mechanisms.

\subsection{Segmentation Method}
We formulate segmentation pretraining as a dense prediction task to explicitly inject pixel-level anatomical supervision into the representation learning process for retinal OCT images. Given an input B-scan image ($x\in R^{H\times W}$), the model predicts a pixel-wise semantic label map ( $y\in0,1,\ldots,C-1^{H\times W}$). Unlike image-level pretraining objectives, this formulation encourages the encoder to preserve spatially grounded and anatomically aligned feature representations, which are critical for downstream classification and explanation tasks.
We adopt a convolutional encoder–decoder architecture implemented using the Segmentation Models PyTorch (SMP) \cite{Iakubovskii2019} framework. The model follows a U-shaped design with an encoder that extracts hierarchical features through progressive downsampling and a symmetric decoder that restores spatial resolution via upsampling and skip connections. While transformer-based segmentation models have demonstrated strong performance in medical imaging, we intentionally employ convolutional architectures to preserve strict spatial locality and stable feature–pixel correspondence, which are essential for reliable downstream attribution analysis.
Two encoder backbones are considered. ResNet-50 \cite{resnet} is selected as a residual-based architecture with strong optimization stability and extensive validation in medical image analysis, while EfficientNet-B4 \cite{effnet} serves as a parameter-efficient alternative based on compound scaling. Both encoders are initialized with ImageNet-pretrained weights and coupled with an identical decoder configuration, ensuring that observed differences in learned representations arise from encoder design rather than decoder capacity.

The segmentation model is trained using a composite loss function that combines region-level overlap accuracy and pixel-wise classification fidelity:
\begin{equation}
\mathcal{L} = \mathcal{L}_{\mathrm{Dice}} + \lambda \mathcal{L}_{\mathrm{CE}},
\end{equation}
where $\mathcal{L}_{\mathrm{Dice}}$ denotes the multi-class Dice loss and $\mathcal{L}_{\mathrm{CE}}$ denotes the pixel-wise cross-entropy loss. We set $\lambda = 1$ in all experiments. Dice loss emphasizes region-level anatomical consistency and boundary alignment, while cross-entropy stabilizes optimization at the pixel level. This combination effectively mitigates class imbalance arising from large variations in retinal layer thickness while maintaining reliable convergence behavior.

Overall, segmentation pretraining enforces spatial correspondence throughout the encoder hierarchy, yielding representations that remain aligned with anatomically meaningful retinal structures. This property is essential for downstream explanation methods that assume a meaningful correspondence between internal feature activations and localized image evidence.

\subsection{Segmentation Experiment}

\subsubsection{Datasets}
Segmentation pretraining is performed using a combined set of NR206 and Duke DME, which together cover both normal retinal anatomy and diabetic macular edema–related pathological variation. Duke DME provides pixel-level annotations for retinal layers and fluid regions and is widely used as a benchmark for OCT segmentation, while NR206 contributes additional anatomical diversity in healthy eyes.

To ensure semantic consistency across datasets, we adopt the Duke DME annotation protocol as the reference label space. Its upper and lower background regions are merged into a single background class, and the labels of the NR206 labels are mapped to this harmonized scheme. After label harmonization, the final segmentation task contains $C = 9$ classes, corresponding to eight retinal layers or regions and one background class. This procedure enforces consistent anatomical semantics across datasets and reduces the risk that the encoder learns dataset-specific annotation artifacts.

We follow the original training/validation partitions provided by each dataset and do not introduce cross-dataset mixing between splits. Additionally, there is no subject overlap within or across splits.

\subsubsection{Training Protocol}
All images are resized to a fixed input resolution before training. We apply data augmentation during training to improve robustness while preserving anatomical plausibility. Augmentations include small geometric transformations and intensity perturbations constrained to physiologically plausible ranges, while preserving the canonical top–down retinal orientation of OCT B-scans.
Models are trained with Adam optimizer using the same learning-rate schedule across experiments. Batch size, learning rate, and training duration are held constant across encoder backbones to isolate the effect of encoder architecture. Training proceeds until validation convergence, with early stopping based on segmentation performance. All experiments are implemented in PyTorch using SMP.

\subsubsection{Evaluation \& Results}
Segmentation performance is evaluated using Intersection over Union (IoU) and Dice coefficient, averaged across all classes. Quantitative results are summarized in Table~\ref{tab:seg_results}.

\begin{table}[htbp]
\centering
\caption{Retinal OCT segmentation performance with different encoder backbones.}
\label{tab:seg_results}
\setlength{\tabcolsep}{6pt}
\renewcommand{\arraystretch}{0.95}
\small
\begin{tabular}{lcc}
\hline
\textbf{Encoder Backbone} & \textbf{IoU} $\uparrow$ & \textbf{Dice} $\uparrow$ \\
\hline
ResNet-50          & 0.8313 & 0.9015 \\
EfficientNet-B4    & 0.7987 & 0.8670 \\
\hline
\end{tabular}
\end{table}

Both encoder backbones achieve strong segmentation performance, consistent with prior work on convolutional OCT segmentation. ResNet-50 attains higher overlap-based metrics, while EfficientNet-B4 delivers competitive performance with a lower parameter count. Given that the objective of this stage is representation initialization rather than optimal segmentation accuracy, both models provide sufficiently strong anatomical supervision.

After segmentation pretraining, the encoder weights are reused to initialize downstream classifiers. By training under explicit anatomical supervision, the encoders are biased toward structure-aware representations that emphasize retinal layer geometry rather than purely task-specific discriminative patterns. This structured initialization encourages downstream decision functions and attribution maps that focus on clinically meaningful anatomical regions, improving interpretability without adding complexity.

\section{Case Studies on UF Cohort}
\label{app:full_case_study}
\begin{figure*}[htbp]
    \centering
    \includegraphics[width=\linewidth]{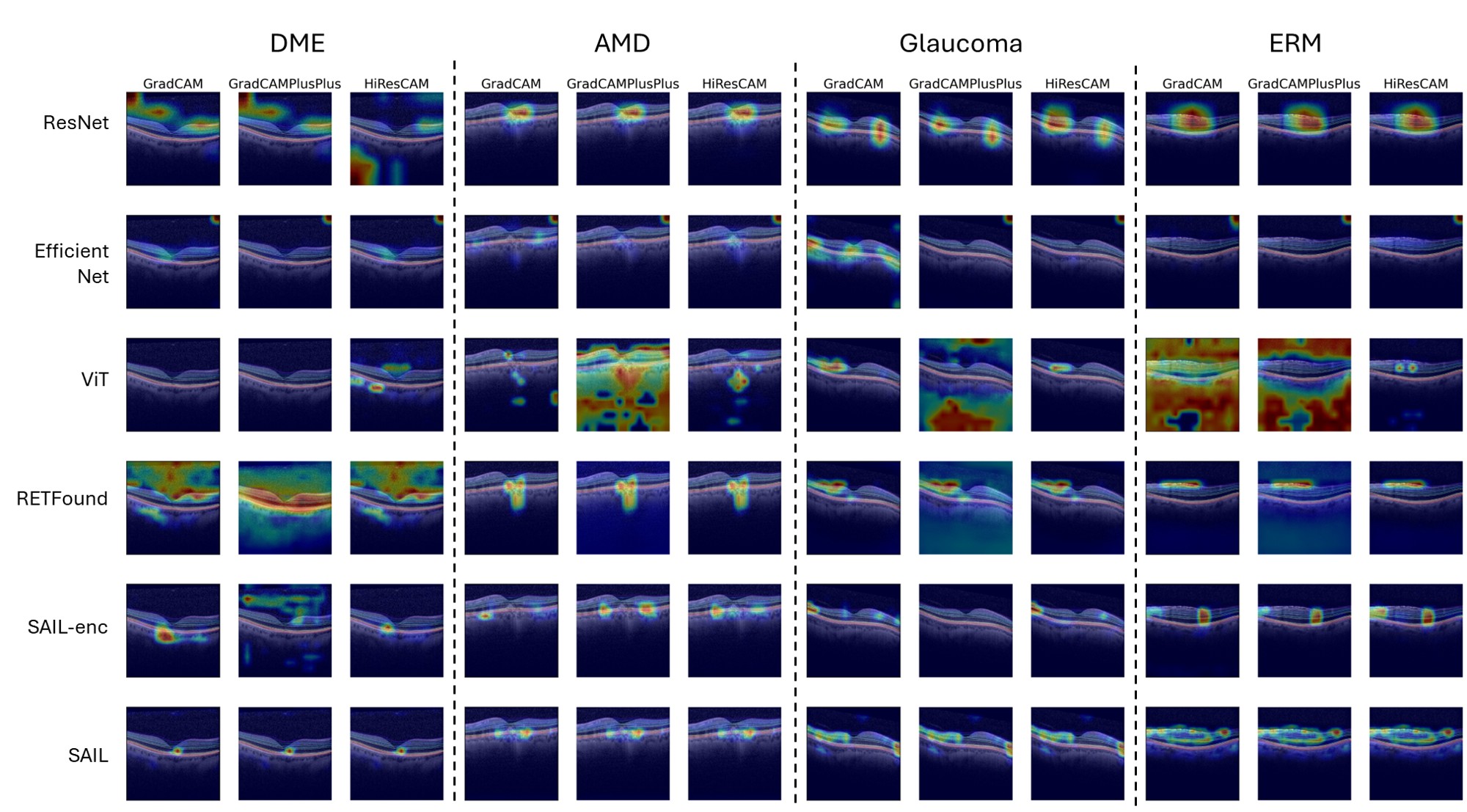}
    \caption{Case-study attribution map comparisons across diseases, models, and XAI methods.} 
    \label{fig:casestudy2}
\end{figure*}

Figure~\ref{fig:casestudy2} presents representative B-scans for four tasks (DME, AMD, Glaucoma, ERM). All attribution maps use the same colormap and normalization for direct comparison. Across diseases and Grad-CAM variants, SAIL produces the most anatomically grounded explanations: activations are largely confined to retinal tissue with minimal background/artifact leakage, consistent with its top RMA/RRA in Figure 3. In contrast, baselines are less stable: RETFound and ResNet-50 exhibit broader, noisier attention and background contamination; ViT and EfficientNet are worst in these examples, frequently dominated by non-diagnostic background signals.
For each disease, SAIL tends to form coherent, localized responses within the retina (often mid-to-lower layers for DME and AMD), while Glaucoma and ERM cases also show more inner-to-mid retinal focus. Overall, these case studies support that structural priors improve specificity and anatomical validity.

\section{Classification Performance with Complete Metrics}
\label{app:full_class_perf}

Tables~\ref{tab:uf_full_cls_perf}~and~\ref{tab:public_full_cls_perf} report the full classification results on the UF cohort and public OCT benchmarks. Overall, SAIL and SAIL-enc achieve performance comparable to, and in several settings better than, strong baselines (ResNet, EfficientNet, ViT, and RETFound) across AUROC, AUPRC, and other standard classification metrics. These results suggest that the proposed structure-aware learning improves interpretability without sacrificing predictive accuracy.

\begin{table*}[!htbp]
\centering
\caption{Full classification performance (\%) on UF tasks.}
\label{tab:uf_full_cls_perf}
\setlength{\tabcolsep}{3.5pt}
\renewcommand{\arraystretch}{0.95}
\small
\begin{tabular}{llcccccc}
\toprule
\textbf{Task} & \textbf{Metric} & \textbf{ResNet} & \textbf{EfficientNet} & \textbf{ViT} & \textbf{RETFound} & \textbf{SAIL-enc} & \textbf{SAIL} \\
\midrule
\multirow{7}{*}{UF DME}
& AUROC     & 96.8 & 95.5 & 96.1 & 98.1 & 97.7 & 97.8 \\
& AUPRC     & 96.0 & 92.7 & 95.7 & 98.0 & 97.2 & 97.4 \\
& Accuracy  & 91.9 & 92.3 & 91.1 & 93.2 & 93.9 & 93.6 \\
& Precision & 91.2 & 91.6 & 90.4 & 92.5 & 93.4 & 92.9 \\
& Recall    & 91.8 & 92.4 & 90.9 & 93.7 & 93.8 & 93.8 \\
& F1        & 91.5 & 92.0 & 90.6 & 93.0 & 93.6 & 93.3 \\
& Kappa     & 83.0 & 83.9 & 81.3 & 85.9 & 87.2 & 86.6 \\
\midrule
\multirow{7}{*}{UF AMD}
& AUROC     & 94.6 & 95.1 & 95.9 & 97.3 & 96.8 & 96.6 \\
& AUPRC     & 91.9 & 92.8 & 95.2 & 97.1 & 96.3 & 96.0 \\
& Accuracy  & 90.4 & 90.8 & 89.6 & 91.3 & 90.7 & 90.8 \\
& Precision & 89.5 & 89.9 & 88.7 & 90.4 & 89.7 & 89.9 \\
& Recall    & 90.3 & 90.5 & 89.4 & 91.3 & 91.0 & 90.5 \\
& F1        & 89.9 & 90.2 & 89.0 & 90.8 & 90.2 & 90.2 \\
& Kappa     & 79.7 & 80.4 & 78.1 & 81.7 & 80.5 & 80.4 \\
\midrule
\multirow{7}{*}{UF Glaucoma}
& AUROC     & 89.6 & 90.1 & 92.2 & 94.1 & 93.6 & 93.5 \\
& AUPRC     & 87.3 & 87.7 & 92.0 & 93.6 & 93.3 & 93.2 \\
& Accuracy  & 83.3 & 84.6 & 84.2 & 86.8 & 86.0 & 85.8 \\
& Precision & 83.1 & 84.5 & 84.0 & 87.0 & 86.1 & 85.7 \\
& Recall    & 83.2 & 84.3 & 84.0 & 86.3 & 85.5 & 85.6 \\
& F1        & 83.1 & 84.4 & 84.0 & 86.5 & 85.7 & 85.6 \\
& Kappa     & 66.3 & 68.8 & 68.0 & 73.1 & 71.5 & 71.3 \\
\midrule
\multirow{7}{*}{UF ERM}
& AUROC     & 94.3 & 91.3 & 94.3 & 95.6 & 95.2 & 95.3 \\
& AUPRC     & 93.9 & 88.8 & 94.1 & 95.5 & 95.2 & 95.1 \\
& Accuracy  & 87.9 & 87.0 & 86.2 & 89.0 & 88.8 & 88.8 \\
& Precision & 87.8 & 86.9 & 87.0 & 89.0 & 88.8 & 88.7 \\
& Recall    & 87.8 & 87.1 & 85.7 & 89.0 & 88.7 & 88.7 \\
& F1        & 87.8 & 86.9 & 86.0 & 89.0 & 88.8 & 88.7 \\
& Kappa     & 75.6 & 73.9 & 72.1 & 77.9 & 77.5 & 77.4 \\
\bottomrule
\end{tabular}
\end{table*}


\begin{table*}[!htbp]
\centering
\caption{Full classification performance (\%) on public datasets.}
\label{tab:public_full_cls_perf}
\setlength{\tabcolsep}{3.5pt}
\renewcommand{\arraystretch}{0.95}
\small
\begin{tabular}{llcccccc}
\toprule
\textbf{Task} & \textbf{Metric} & \textbf{ResNet} & \textbf{EfficientNet} & \textbf{ViT} & \textbf{RETFound} & \textbf{SAIL-enc} & \textbf{SAIL} \\
\midrule

\multirow{7}{*}{OCTDL AMD}
& AUROC     & 99.7 & 99.0 & 99.1 & 99.6 & 99.7 & 99.8 \\
& AUPRC     & 99.4 & 98.1 & 98.0 & 99.1 & 99.2 & 99.4 \\
& Accuracy  & 98.1 & 86.6 & 93.3 & 98.7 & 96.5 & 98.7 \\
& Precision & 96.3 & 92.7 & 88.1 & 97.6 & 93.2 & 97.2 \\
& Recall    & 98.2 & 68.7 & 95.7 & 98.6 & 97.2 & 99.2 \\
& F1        & 97.2 & 73.3 & 91.0 & 98.1 & 95.0 & 98.1 \\
& Kappa     & 94.4 & 48.4 & 82.1 & 96.2 & 90.1 & 96.3 \\
\midrule

\multirow{7}{*}{OCTDL DME}
& AUROC     & 99.5 & 97.5 & 99.6 & 100.0 & 99.2 & 100.0 \\
& AUPRC     & 99.5 & 97.6 & 99.5 & 100.0 & 99.1 & 100.0 \\
& Accuracy  & 97.9 & 95.9 & 96.9 & 97.9 & 97.9 & 97.9 \\
& Precision & 98.6 & 96.0 & 97.9 & 98.6 & 98.6 & 98.6 \\
& Recall    & 96.7 & 94.3 & 95.0 & 96.7 & 96.7 & 96.7 \\
& F1        & 97.5 & 95.1 & 96.3 & 97.5 & 97.5 & 97.5 \\
& Kappa     & 95.1 & 90.2 & 92.6 & 95.1 & 95.1 & 95.1 \\
\midrule

\multirow{7}{*}{OCTDL ERM}
& AUROC     & 99.7 & 91.9 & 100.0 & 99.3 & 99.9 & 99.3 \\
& AUPRC     & 99.6 & 91.1 & 100.0 & 99.2 & 99.9 & 99.2 \\
& Accuracy  & 95.9 & 61.2 & 100.0 & 95.9 & 96.9 & 94.9 \\
& Precision & 97.2 & 69.3 & 100.0 & 97.2 & 97.9 & 96.5 \\
& Recall    & 93.5 & 69.9 & 100.0 & 93.5 & 95.2 & 91.9 \\
& F1        & 95.1 & 61.2 & 100.0 & 95.1 & 96.4 & 93.8 \\
& Kappa     & 90.2 & 30.8 & 100.0 & 90.2 & 92.7 & 87.7 \\
\midrule

\multirow{7}{*}{OCT2017 DME}
& AUROC     & 100.0 & 100.0 & 100.0 & 100.0 & 100.0 & 100.0 \\
& AUPRC     & 100.0 & 100.0 & 100.0 & 100.0 & 100.0 & 100.0 \\
& Accuracy  & 99.6  & 99.4  & 100.0 & 99.6  & 99.6  & 99.8 \\
& Precision & 99.6  & 99.4  & 100.0 & 99.6  & 99.6  & 99.8 \\
& Recall    & 99.6  & 99.4  & 100.0 & 99.6  & 99.6  & 99.8 \\
& F1        & 99.6  & 99.4  & 100.0 & 99.6  & 99.6  & 99.8 \\
& Kappa     & 99.2  & 98.8  & 100.0 & 99.2  & 99.2  & 99.6 \\
\bottomrule
\end{tabular}
\end{table*}

\end{document}